\documentclass[twoside]{article}

\usepackage[preprint]{aistats2027}
\usepackage[hidelinks]{hyperref}
\usepackage{url}
\usepackage{graphicx}
\usepackage[makeroom]{cancel}

\usepackage{amsmath,amsthm,amsfonts,bm,dsfont}
\usepackage[
lined,
ruled,
vlined]{algorithm2e}
\usepackage[
capitalise,
noabbrev,
]{cleveref}
\AddToHook{cmd/appendix/before}{%
    \crefalias{section}{appendix}%
    \crefalias{subsection}{appendix}%
    \crefalias{subsubsection}{appendix}
}
\usepackage{./raccourcitron}
\usepackage{booktabs}
\usepackage{subcaption}
\usepackage{float}
\usepackage{xcolor}

\DeclareMathOperator{\kvol}{kVol}
\DeclareMathOperator{\vol}{Vol}

\newtheorem{thm}{Theorem}

\theoremstyle{definition}
\newtheorem{definition}{Definition}

\theoremstyle{remark}
\newtheorem*{remark}{Remark}

\newcommand{\kvv}{z} 
\newcommand{\KVV}{Z} 
\newcommand{\Dlab}{\mathcal{D}^{\mathrm{lab}}}
\newcommand{\Qlab}{\mathcal{Q}^{\mathrm{lab}}}
\newcommand{\Dpool}{\mathcal{D}^{\mathrm{pool}}}
\newcommand{\Dtrain}{\mathcal{D}^{\mathrm{tr}}}

\usepackage[
natbib,
backend=biber,
style=numeric,
sorting=nyt,
backref=false,
maxcitenames=2,
maxbibnames=10,
minbibnames=4,
url=false,
eprint=false,
]{biblatex}

\newcommand{\paren}[1]{\left( #1 \right)}

\newcommand{\acc}[1]{\left\{ #1 \right\}}

\begin{document}

%

%

\twocolumn[

\aistatstitle{$k$-Volume: a Unified Active Learning Strategy}

\aistatsauthor{ Eliot Tron \And Alain Celisse \And  Christophe Denis \And Alice Le Brigant }

\aistatsaddress{SAMM Université Paris 1, \\ IRIT, INP Toulouse \And SAMM, Université Paris 1 \And SAMM, Université Paris 1 \And SAMM, Université Paris 1} 

]


\begin{abstract}
%
%
%

In the classification context where revealing the label of a sample is costly, we propose a new  
Active Learning (AL) strategy named \emph{$k$-Volume (kVol)}. %
It randomly
selects a sample to be labeled according to the volume
of the smallest ball containing its $k$ nearest neighbors. 
kVol is highly versatile: it can be applied either in the input space or in the output space, and it enjoys the desirable properties of the three main families of AL strategies: density, diversity, and (when performed in the output space) uncertainty. 
From a theoretical perspective, we establish that kVol is asymptotically equivalent to uniform sampling over the support of the (input or output) data distribution.
%
%
%
%
%
Based on experiments on various synthetic and real-life tabular datasets and images, 
we observe that 
kVol is surprisingly stable: whatever the classifier or the experimental setup, it still lies among the best AL strategies unlike other competitors that fail dramatically. 
In particular, kVol outperforms state-of-the-art competitors such as \emph{CoreSet}, \emph{TypiClust} or \emph{Uncertainty} on several real-life datasets, at a low computational cost.

%
%
%
\end{abstract}

\section{Introduction}
\label{sec:intro}



Training a classifier requires labeled data, which is a bottleneck in situations where accessing the labels costs significant human resources or intensive computing \citep{casilliAttendantRobotsEnquete2019}. For medical image labeling for example, doctors have to examine images one-by-one, which is highly time consuming.
Active learning (AL) \citep{settlesActiveLearningLiterature2009, liSurveyDeepActive2024} has been introduced to tackle this limitation: many unlabeled data are available, however the labeling process of such data is expensive.

AL strategies rely on \emph{query functions} designed to iteratively return the ``best'' (in a sense that needs to be defined) samples to label.
The way query functions are crafted can be categorized based on their objective. For instance influence-based strategies choose samples with the greatest impact on the loss \citep{wangDeepActiveLearning2022}. 
The present work focuses on the three main families of AL strategies:

\textbf{Density family.} Strategies of this family select samples that are most representative of the entire unlabeled set distribution. 
    The \emph{Uniform} AL strategy belongs to this family since sampling uniformly over the available samples amounts to estimating the distribution (density) of these samples.
    More sophisticated strategies of this family involve weights based on the similarity to points in the unlabeled dataset \citep[Sec. 3.6]{settlesActiveLearningLiterature2009}. 
    However strategies of the density family struggle in the imbalanced case \citep{chenSurveyImbalancedLearning2024, gaoComprehensiveSurveyImbalanced2026}. In this case, these strategies miss out the underrepresented labels.

\textbf{Diversity family.} These strategies select samples that differ the most from the ones that have been already labeled, which is exactly the behavior of the so-called \emph{Diversity} AL strategy.
    The query functions aim at exploring the data space \citep{krauseNonmyopicActiveLearning2007, elreedyNovelActiveLearning2019}.
    These strategies typically select points that are far from each other, thereby avoiding getting stuck in very narrow regions of the input space. However, such strategies can struggle with high-variance classes.
    It should be noted that neither the density family, nor the diversity family takes into account the classifier local performance. Then any AL strategy from both these families can ignore regions where the suboptimal performance of the classifier would require more observations.

\textbf{Uncertainty family.} Strategies of this family (for instance the so-called \emph{Uncertainty} AL strategy) select samples for which the classifier has the weakest confidence.
    Unlike the density and diversity families, the uncertainty one is precisely designed to incorporate information about the running classifier. 
    For instance, the confidence of the classifier is measured by means of the entropy criterion \citep{settlesActiveLearningLiterature2009, wangCostEffectiveActiveLearning2017}, or more sophisticated criteria based on the Fisher information or variants \citep{kirschUnifyingApproachesActive2022, gulFisherMaskEnhancingNeural2024}.
    These strategies suffer from mainly focusing on the neighborhood of the classifier boundaries. By contrast, they overlook significant variations in unexplored regions (see also \cref{app:shortcoming_uncertainty_AL} and \cref{fig:counter_example_proba_based_al}).

While each of the above families comes with its own range of situations where it outperforms the others, no AL strategies from any of these families perform uniformly better than the others in all cases \citep{munjalRobustReproducibleActive2022,eisenhardtMindGapFramework2026, huExploringLimitationsActive2021,popDeepEnsembleBayesian2018}. 
To the best of our knowledge, a state-of-the-art AL strategy improves on the previous ones by combining several of them from different families. 
%
%
One main achievement of the present work is to introduce a new AL strategy called \emph{$k$-Volume}, which avoids the caveats of each individual families by combining ideas from several of them.

\paragraph*{Related works.}
%
The literature has already explored the idea of combining strategies from several families.
A first example consists in incorporating diversity in a predefined query function by randomizing some of the returned samples as suggested 
in \citet{bouneffoufContextualBanditActive2014,bouneffoufExponentiatedGradientExploration2016}. 
There, a random choice is made at each step between either using the query function or uniformly selecting a sample among the remaining ones. 
More generally, examples rely on a weighted sum of query functions from different families \citep{xuIncorporatingDiversityDensity2007, cebronActiveLearningObject2009,keeQuerybycommitteeImprovementDiversity2018}.
In both cases, the main limitation owes to finding the weights, which can be a tedious task (it can require multiple parameters optimization every time a new task is considered).

Other strategies combine features of the diversity and density families in a more intrinsic way (without requiring the need for parameter tuning). 
The \emph{CoreSet} strategy \citep{senerActiveLearningConvolutional2018} selects a subset of the samples such that the union of balls centered at them yields a covering of the entire dataset. 
%
%
Another example is \emph{TypiClust} \citep{hacohenActiveLearningBudget2022}, which is based on a notion of ``typicality'' (it computes the inverse of the average Euclidean distance of a point to its neighbors (density) within a precomputed cluster (diversity)).
However one strong limitation of these two strategies comes from their high computational costs. Computing the clustering (TypiClust) or designing a covering (CoreSet) are time-consuming tasks, which become even prohibitive as the dimension $d$ of the input space gets large in practical situations.

\paragraph*{Main contributions.}
In this paper, a new active learning (AL) strategy called \emph{$k$-Volume (kVol)} is introduced, which randomly selects a sample to be labeled according to the volume of the smallest ball containing its $k$ nearest neighbors. 
Applied on the input space, the kVol strategy (kVolIn) combines the objective of the density family (random sampling) with the one of the diversity family (uniform exploration of the space). 
This interpretation is theoretically justified: kVol is asymptotically equivalent to \emph{uniform sampling over the support} of the data distribution.
The kVol strategy is \emph{highly versatile} since it can also be applied on the output space (kVolOut), which allows for incorporating the uncertainty-objective as well. 
By means of empirical experiments on synthetic and real-life (tabular and image-based) datasets, three main strengths of kVol are emphasized: $(i)$ Unlike other competitors, it is \emph{remarkably stable} since it always stands among the best strategies no matter the context, $(ii)$ It \emph{outperforms state-of-the-art strategies} on several real-life datasets, and $(iii)$ It is \emph{computationally cheap} compared to TypiClust or CoreSet.


\section{Background and Problem Statement}
\label{sec:back_pb_statement}

This section introduces the multi-class classification framework as well as the active learning setting. 
\paragraph*{Multi-class classification.}
Let us consider a set of independent and identically distributed couples $\Dtrain=(X_1,Y_1),\ldots,(X_n,Y_n) \in\mathcal{X} \times \mathcal{Y}$ with the unknown probability distribution $P$, where $\mathcal{X}$ denotes the input space and the output space is $\mathcal{Y} = [C]:= \{1,\ldots,C\}$, for $C \geq 2$.  
A classifier is a measurable map $f \in \mathcal{F} = \acc{ h: \mathcal{X} \to \mathcal{Y} }$. 
The \emph{passive} classification setup assumes that for the data in 
$\Dtrain$ the labels $Y_i$'s are all available. This allows to optimize an empirical criterion based on $\Dtrain$, then leading to a classifier $\hat{f}$ \emph{trained} from $\Dtrain$. 
%

\paragraph*{Pool-based active learning.}

Unlike the above \emph{passive} learning framework, Active Learning does not allow for accessing all the labels in $\Dtrain$.
This setting accounts for the realistic situation where revealing any label is highly costly or prohibited. 
By contrast, the AL setting only gives access to a limited number $B\ll n$ of labels that is called the \emph{total budget}. 
The present work focuses on the \emph{pool-based} AL setup, where a pool of \emph{unlabeled} data $\Dpool = \acc{X_1, \ldots, X_n}$ is available.

If $\cD\subset \Dpool$ 
denotes the input points whose labels have already been revealed and $\Dlab = \acc{(X_i,Y_i) \mid 1\leq i\leq n, X_i \in \cD}$ the corresponding labeled data, then the classifier is trained from $\Dlab$ leading to $\hat f_{\Dlab}$. 
The purpose of any AL algorithm is then to query the labels of points within $\Dpool$ (up to the total budget $B$) by following a dedicated query rule such that the resulting classifier $\hat f_{\Dlab}$ is \emph{optimal}. 
Here the optimality consists in minimizing the prediction error 
\begin{align}\label{eq.prediuction.error}
    PE( f ) = \E_{\pa{X,Y}\sim P}\bra{\ell\pa{ f(X), Y}},
\end{align}
where $\pa{X,Y}\sim P$ is independent of $\Dtrain$, and $\ell\paren{\cdot,\cdot}: \mathcal{Y} \times \mathcal{Y} \to \R_+$ denotes a loss function. Examples of such loss functions include the 0-1--loss and the soft-max loss, to name but a few.

\paragraph{Ideal strategy.}
In this context, a naive intuition could suggest that one can access an optimal set $\cD^\star \subset \Dpool$ with cardinality $B$ such that the prediction error~\eqref{eq.prediuction.error} is minimized. It turns out that such an optimal set $\cD^\star$ would solve the following optimization problem
 \begin{equation}\label{eq:al_def_min_loss}
				\cD^* \in \argmin_{\substack{\cD \subset \Dpool\\ \abs{\cD} = B}} PE\paren{ \hat{f}_{\Dlab}}.
\end{equation}    
However uncovering $\cD^\star$ would require varying $\cD$ within $\Dpool$ and since each set $\cD$ has cardinality $B$, this would clearly violate the AL constraint that at most $B$ labels can be revealed.
Let us notice that another reason for \cref{eq:al_def_min_loss} to be difficult to solve owes to the expression of the criterion to be optimized. It requires $\hat{f}_{\Dlab}$ to be trained for any candidate set $\cD$ before deciding whether the corresponding $\Dlab$ (containing revealed labels) is a relevant candidate. Each time $\Dlab$ is finally discarded, the corresponding revealed labels are finally wasted.

\paragraph{Overcoming these two difficulties.}
To overcome these issues, two ingredients are then combined.
The first one is a common paradigm called \emph{batch} AL \citep{citovsky2021batch}, where the unique query step of $B$ points in $\Dlab$ is replaced by the successive queries of $T\in\N$ \emph{batches} $\cQ_t$, with $t\in [0,T]\cap\N$, each with cardinality $b < B$ (or $b_0$ for the initialization, see Algorithm~\ref{algo:AL algorithm.}).
%
Then, each time a batch $\cQ_t$ of $b$ points has been queried and the corresponding labels revealed, the labeled data $Q_t^{\rm lab}$ are combined with the previously available ones $\cD^{\rm lab}_{t-1}$. The classifier is trained from $\cD^{\rm lab}_{t}$, and the process is iterated until the total budget of $B$ labels has been consumed.
The second ingredient is the use of a \emph{proxy} objective function instead of the prediction error~\eqref{eq.prediuction.error} itself. 
It involves selecting the next samples to be labeled based on a \emph{query function} denoted $\alpha$.
Examples of such query functions are detailed below, each being chosen among the leading families: 
density, diversity, or uncertainty.
%
%
%
\begin{algorithm}
	\DontPrintSemicolon
	\caption{(Batched) Active Learning.}
	\label{algo:AL algorithm.}
	\KwIn{Unlabeled Data $\Dpool$.}
	\KwData{Initial batch size $b_0$, step batch size $b$, max iterations $T$, max labeling budget $B$, query function $\alpha$, trainable model $\hat{f}$.}
  $\cQ_0 \leftarrow$ Initial sampling from $\Dpool$ where $\abs{\cQ_0} = b_0$\;
  $\Qlab_0 \leftarrow$ Labeled version of $\cQ_0$\;
  	$\Dlab_0 \leftarrow \Qlab_0$ \tcp*{Initialization of trainset.}
    $\hat{f}_{\Dlab_0} \leftarrow \hat{f} \text{ trained on } \cD_0^{\text{lab}}$\;
	$t \gets 0$ \;
	\While{$\abs{\cD_t^{\text{lab}}} < B$ \textbf{\textup{and}} $t < T$}{
    $t \gets t + 1$ \;
    $\cQ_t \leftarrow \alpha\pa{\hat{f}_{\Dlab_{t-1}}, \Dpool_{t-1}, b}$ \tcp*{Query set.}
    $\Qlab_t \leftarrow$ Labeled version of $\cQ_t$\;
    $\Dlab_t \leftarrow \Dlab_{t-1} \cup \Qlab_t$\;
    $\hat{f}_{\Dlab_t} \leftarrow \hat{f}_{\Dlab_{t-1}} \text{ updated from } \cD_t^{\text{lab}}$\;					$\Dpool_t \leftarrow \Dpool_{t-1} \setminus \cQ_t$\;
	}
	\KwOut{The last trained model $\hat{f}$.}
\end{algorithm}

\paragraph{Query functions.}
Let $t\in\mathbb{N}$ denote the running iteration number, and $\Dpool_t$ the set of samples remaining available at iteration $t$ of the process. In particular with $t\geq 1$, $\Dpool_t = \Dpool_{t-1} \setminus \cQ_{t} $, where $\cQ_{t}$ denotes the set of points for which labels have been revealed at step $t$.  
See \cref{algo:AL algorithm.} which further describes the general structure of any (batch) AL strategy.
Hereafter, we provide a few examples of query functions.

\emph{Uniform query function:} For any $t\geq 1$,
\begin{align*}
    \alpha_{\text{uniform}}\pa{\hat{f}_{\Dlab_{t-1}}, \Dpool_{t-1}, b} = \set{x_1, \ldots, x_b} , 
\end{align*}
where $x_1,\ldots,x_b$ are uniformly sampled (without replacement) from $\Dpool_{t-1}$.
This query function belongs to the density family. It will serve as a baseline in the experiments exposed in \cref{sec:num_results}.

\emph{Score-based query functions:} 
Such query functions select the $b$ points according to a score $s$ that is user-specified (see examples below). 
Then, after ordering the samples in $\Dpool_{t-1}$ according to their scores as $s\pa{x_{(1)}} \geq \ldots \geq s\pa{x_{\pa{\abs{\Dpool_{t-1} }}}}$, it results
\begin{align*}
    \alpha_{s}\pa{\hat{f}_{\Dlab_{t-1}}, \Dpool_{t-1}, b} = \set{x_{\pa{1}}, \ldots, x_{\pa{b}}}.
\end{align*}   
Classical examples of such scores are 
\begin{align*}
s(x) &= \min_{(x^\prime, y^\prime) \in \Dlab_{t-1}}d(x, x^\prime),\tag{Diversity}\\
s(x) &= 1 - \max_{1\leq j\leq C} \hat{p}_j(x), \tag{Uncertainty}\\
s(x) &= - \sum_{j=1}^C \hat{p}_j(x) \log \hat{p}_j(x), \tag{Entropy}
\end{align*}
where $d(\cdot,\cdot)$ is any (pseudo-)distance defined on $\mathcal{X}$ and $\hat p(x) = (\hat p_1(x),\ldots,\hat p_C(x)) \in [0,1]^C$ such that each $\hat p_j(x) $ is an approximation to $\P\pa{Y=j\mid X=x}$.

        
%
%
As discussed in the introduction, each family of query functions has its own set of situations where it outperforms the others. However practical situations emphasize that the best performances are achieved by functions leveraging the merits of several families (see the examples mentioned along the Related Works section). 

\section{The $k$-Volume Query Function}
\label{sec:our_method}

\begin{figure}[h]
    \centering
    \includegraphics[width=0.4\linewidth]{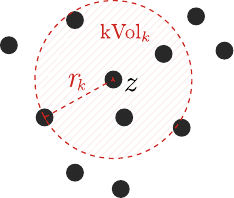}
    \caption{Illustration of $\kvol_k$ with $k=4$.}
    \label{fig:kVol_basics}
\end{figure}

The present section describes a new AL strategy based on the computation, for a given point, of the volume of the ball centered at that point and whose radius is the distance to its $k$th nearest neighbor.
This justifies the name of this new AL strategy, that is \emph{$k$-Volume}.
\begin{definition}
    For any $z$ in a metric space $(\mathcal{Z},d(\cdot,\cdot))$ with $\mathcal{Z} \subset \R^d$, let $\cD \subset \mathcal{Z}$ be a finite set.
    For any $1\leq k\leq \abs{\cD}-1$, the $k$-volume function $\kvol_k$ is defined by
    \begin{equation*}
        \kvol_k(\kvv, \cD) = \vol\pa{\cB\pa{\kvv, r_k(\kvv)}},
    \end{equation*}
    where 
        $r_k(\kvv) := \min \big\{r \geq 0 \,|\, \exists S \subset \cB(\kvv,r)\cap\cD, \text{ with } z \not\in S,~ \abs{S} = k\big\}$, and $\vol\pa{\cB\pa{0, 1}}= \pi^{d/2} (\Gamma(d/2+1))^{-1}$.
 %
\end{definition}
The $\kvol_k(\cdot)$ function outputs the volume of the smallest ball centered at $z$ which contains the $k$ nearest neighbors of $z$ among the points in $\cD\setminus \acc{z}$.
An illustration of this can be found on \cref{fig:kVol_basics}.

Having introduced the $\kvol_k(\cdot)$ function, let us now explain how it connects to the query function of the new $k$-Volume AL strategy.
\begin{definition}\label{def.KVol.query}
	At step $t\geq 1$ and for any integer $1\leq b\leq B$, the $k$-Volume query function $\alpha_{\kvol}$ samples (without replacement) $b$ points in $\Dpool_{t-1}$ with weights given by $\kvol_k(\cdot)$ for $ 1\leq k\leq |\Dpool_{t-1}|-1$, that is
    \begin{gather*}
        \alpha_{\kvol}\pa{\hat{f}_{\Dlab_{t-1}}, \Dpool_{t-1},b} = \set{\kvv_1,...,\kvv_b}, \\
        \intertext{where, for any $1\leq i\leq b-1$,}
        \kvv_{i+1} \mid \kvv_1,...,\kvv_i \sim \sum_{\kvv\in \Dpool_{t-1}\setminus \set{\kvv_1,\ldots,\kvv_i}} \omega_k(z) \delta_{\kvv},
    \end{gather*}
    with 
    \[\omega_k(\kvv) = \frac{\kvol_k(\kvv, \Dpool_{t-1})}{\sum_{\kvv^\prime \in \Dpool_{t-1}\setminus \set{\kvv_1,\ldots,\kvv_i}}  \kvol_k(\kvv^\prime,\Dpool_{t-1})},\] and $\delta_{\kvv}$ stands for the Dirac measure at $\kvv$.
\end{definition}
According to \cref{def.KVol.query}, at each step $t\geq 1$, the query function consists in randomly sampling a batch of $b$ points. Each sample $z\in \Dpool_{t-1}$ is drawn with a probability $\omega_k(z)$ that is proportional to the volume of the ball $\mathcal{B}(z,r_k(z))$, where $r_k(z)$ denotes the distance between $z$ and its $k$th nearest neighbor within $\Dpool_{t-1}$ with respect to $d(\cdot,\cdot)$. Thus at each step, any point $z$ located in \emph{dense} regions of the unlabeled set will have a low weight $\omega_k(z)$ since its distance to its $k$th nearest neighbor is likely to be small. By contrast, low-density regions will induce high sampling weights $\omega_k(z)$ for any $z$ lying there.
Therefore, the $k$-Volume query function is intuitively trying to subsample points from a uniform distribution over the support of the dataset probability distribution, instead of sampling uniformly over the set $\Dpool$. 
This intuition is formalized in the following theorem.
\begin{thm}\label{thm:conv_to_uniform}
    Let $\KVV_1,...,\KVV_n \in\mathcal{Z}\subset \R^d$ be $n$ \textit{i.i.d.} random vectors with probability distribution $\mu$ that is absolutely continuous with respect to the Lebesgue measure, that is $\rho = d\mu/d\lambda$ on $\R^d$.
    Further assume that $\rho$ is uniformly continuous on $\cZ$, $S:=\Supp\mu$ is compact, and $\rho>0$ almost everywhere on $S$.
    If $1\leq k \leq n-1$ is chosen such that
    \begin{align*}
        \frac{k}{\log n}\xra{n\to\infty} +\infty,\quad \mbox{and }\quad  \frac{k}{n} \xra{n\to\infty} 0,
    \end{align*}
    then $\KVV_{I_{k,n}}$ converges in distribution to the uniform distribution on the support of $\mu$, that is 
    \begin{equation*}
        \KVV_{I_{k,n}} \xra[\mathcal{L}]{n\to\infty} \cU\pa{\Supp\mu},
    \end{equation*}
    where $I_{k,n}$ is the index randomly selected by the $k$-Volume query function (see \cref{def.KVol.query} with $b=1$).
\end{thm}
%
%
The proof is deferred to \cref{app:proof_of_conv_to_uniform}. The above convergence holds true as long as $k/\log n \to +\infty$, as $n$ grows. However the $k$-Volume query function described in \cref{def.KVol.query} is applied iteratively with the number $t$ of iterations and as $t$ increases, the cardinality $|\Dpool_t|$ becomes smaller or equal to $n - b t$ (note that Theorem~\ref{thm:conv_to_uniform} only addresses $b=1$). Then the number $k$ of neighbors has to satisfy $k\leq n - bt$. It results the following constraint on the number $T$ of iterations
\begin{align*}
\frac{k}{\log n} \leq   \frac{n - \min(bT,B)}{ \log n}\xra{n\to\infty} +\infty,
\end{align*}
which means that for \cref{thm:conv_to_uniform} to hold, it is necessary to require that $\log n = o\paren{ n - \min(bT,B)}$. Let us emphasize that this requirement is fully consistent with the AL formulation which limits the total number of labels to be revealed to a very small fraction of the total amount of available samples, that is $B\ll n$.

\paragraph{kVolIn and kVolOut.}
From \cref{def.KVol.query}, the choice of the $\kvv \in\mathcal{Z}$ offers a \emph{high versatility} for the $k$-Volume strategy.
Specifically, the space $\mathcal{Z}$ can be set as the input space $\mathcal{X}$ or the output space $\mathcal{Y}$, respectively leading to the kVolIn and kVolOut strategies.
%

    \emph{kVolIn}: Query function from \cref{def.KVol.query} applied to $\kvv := x$ in input space $\mathcal{X}$. Let us emphasize that here, the input space can be any latent space used to embed complex objects such as images for instance. This is precisely what is illustrated in \cref{sub:image_data}.
    
    \emph{kVolOut}: Query function from \cref{def.KVol.query} applied to $\kvv$ in the output space. In \cref{sec:num_results}, the output space is the simplex $\Delta^{C-1}$ and $\kvv = \paren{\hat p_1(x),\ldots,\hat p_{C-1}(x)}$, the $C-1$ first components of $\hat p(x)$, an estimator of the posterior probability vector of each class at location $x\in\mathcal{X}$. 
    However there is no reason to restrict to this setup, and the output space 
    can more generally be any space where the output of a predictor is encoded. 
    
%
It is important to notice that, when applied at the input space level, kVolIn behaves as a query function from both the density and the diversity families. In other words, it combines the assets of both these families: $(i)$ it explores the whole support of the distribution $\mu$ without being influenced by any overrepresented class (unlike strategies from the density family), $(ii)$ it is not limited to explore regions at the boundary of the support (unlike strategies from the diversity family). 

However as long as it is applied at the input space level, kVolIn does not incorporate any information regarding the underlying classifier, which can be seen as a limitation compared to the uncertainty family of AL query functions.
This weak limitation is then alleviated by kVolOut. By working at 
the output space level, it 
incorporates some information about the classifier behavior (for instance by using $\hat p(\cdot)$ as discussed above).
In this respect, kVolOut must be understood as a new AL strategy not only enjoying the assets of the density and diversity families, but also the ones of the uncertainty family. This results in a noticeable improvement compared to the competitors involved in the empirical assessment described in \cref{sec:num_results}.

\section{Numerical Results}
\label{sec:num_results}
In this section, numerical experiments
are conducted on synthetic and real datasets in order to assess the performance of $k$-Volume.
%
%
All accuracy\footnote{Number of correctly classified points divided by the total cardinality of the test set.} curves result from 10 repetitions\footnote{More precisely, seeds ranging from 0 to 9 are used.} of the experimental conditions with different initial seeds.
The plotted curves correspond to average accuracy curves (with the 95\% confidence intervals displayed in the background).
We arbitrarily select a value of $k=4$ for the $k$-Volume strategy in the following experiments. However, \cref{fig:value_for_k_CIFAR10} empirically shows that this value has no significant impact on the statistical performance as long as it remains small compared to the labeling budget (see also \cref{app:value_for_k}). 
The code used to carry out the experiments is available 
at \url{https://codeberg.org/eliot-tron/k-volume-active-learning}.

\subsection{Synthetic data}
\label{sub:synth_data}

Synthetic data are first generated for investigating the behavior of $k$-Volume and its competitors.

\begin{figure}
    \centering
    \begin{subfigure}[c]{0.49\linewidth}
    \centering
    \includegraphics[width=\linewidth]{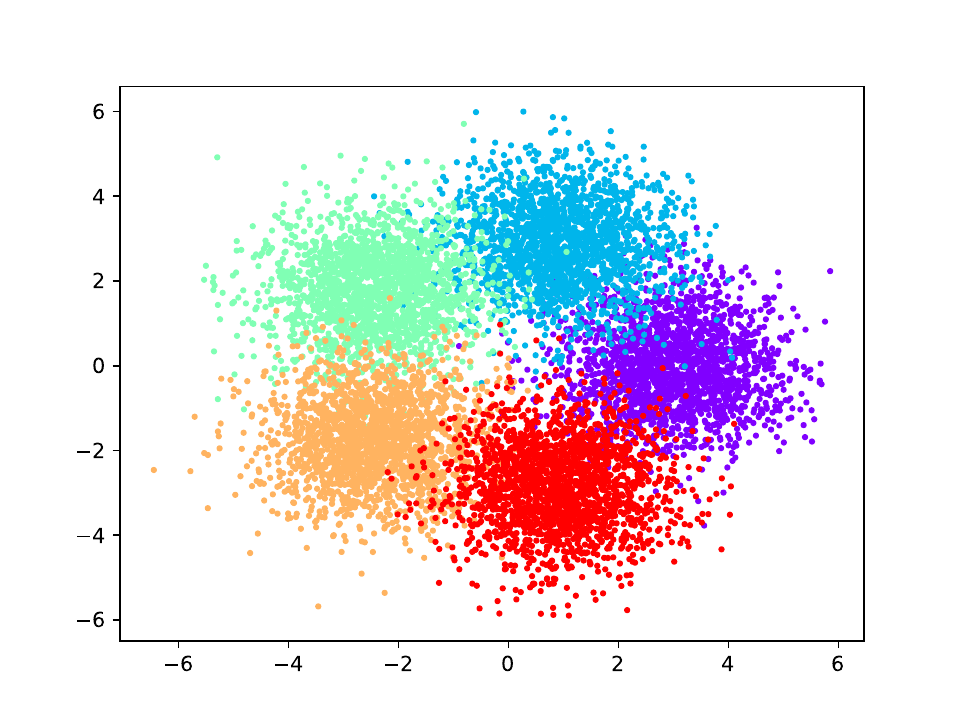}
            \caption{Scenario No. 1}
    \end{subfigure}
    \begin{subfigure}[c]{0.49\linewidth}
    \centering
    \includegraphics[width=\linewidth]{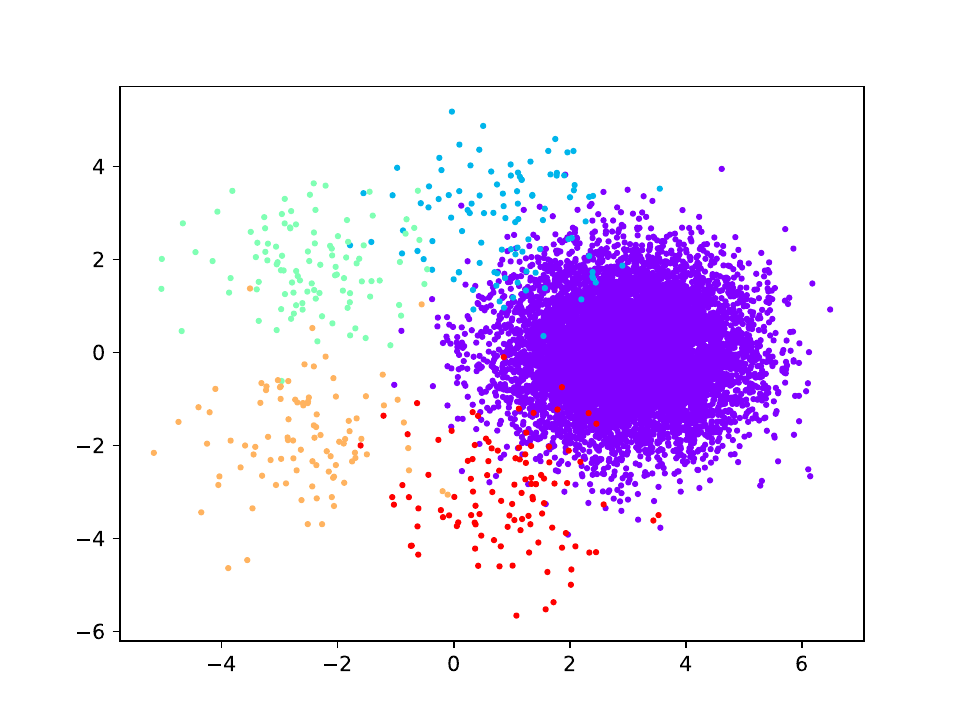}
            \caption{Scenario No. 2}
    \end{subfigure}
    \caption{Visualization of the synthetic training set in the 2D scenarios.}
    \label{fig:synthetic_trainset_viz}
\end{figure}

\paragraph{Simulation design.}
The probability distribution of the \emph{independent} couples $(X_i,Y_i)$, $1\leq i\leq n$, is a Gaussian mixture model with $C=5$ components, that is 
\begin{align*}
\forall i=1,\ldots,n,\qquad & Y_i \sim \mathcal{M}(\pi_1,\ldots,\pi_C) \\
         &   X_i \mid Y_i  \sim \cN\pa{\mu_{Y_i}, \Sigma_{Y_i}},    
\end{align*}
where the $\pi_j$'s are the prior probabilities of each class, $\mathcal{M}(\pi_1,\ldots,\pi_C)$ denotes the induced multinomial distribution, and $\mu_1,\ldots,\mu_C \in \R^d$ (respectively $\Sigma_1,\ldots,\Sigma_C$) are the mean vectors (resp. the covariance matrices) of the different classes.
In the following, the covariance matrices are $\Sigma_1= \dots =\Sigma_C=I_d$ for all classes, with $d\in\acc{2,20}$.
\Cref{tab:synth_exp_params} summarizes 4 scenarios chosen for assessing the impact of both the input dimension $d$ and the class imbalance on the final statistical performance.
\begin{table}[h]
    \caption{Parameters for the numerical experiments on synthetic data.}
    \label{tab:synth_exp_params}
    \centering
    \begin{tabular}{lcccc}
    \toprule
        Scenario & Balanced & Dimension \\ 
    \midrule
         No. 1     & Yes      & 2         \\ 
         No. 2     & No       & 2         \\ 
         No. 3     & Yes      & 20        \\ 
         No. 4     & No       & 20        \\ 
    \bottomrule
    \end{tabular}
\end{table}
In the 2-dimensional scenarios, the classes mean-vectors are located along a circle of radius $3$ with equal angles, that is $\mu_y = \pa{\cos(2\pi y / C), \sin(2\pi y / C)}$ (see \cref{fig:synthetic_trainset_viz}).
For the 20-dimensional scenarios, we replicate the 2-dimensional ones by randomly distributing the mean vectors of each class on a hyper-sphere of radius 3. These mean vectors are sampled in two steps: after being drawn as standard Gaussian vectors, they are rescaled to an $L^2$-norm of $3$ (meaning they no longer perfectly satisfy the equal angles property).
\Cref{fig:synthetic_trainset_viz}, Panel (a) displays the 2-dimensional setup with balanced classes (all prior probabilities are equal to 1/5), while Panel (b) illustrates the situation where one class is strongly dominating the others (imbalanced scenario).
The classifier trained by each AL competitor is the Linear Discriminant Analysis classifier (LDA) (implemented in the Scikit-learn package \citep{pedregosaScikitlearnMachineLearning2011}).
For comparison purposes, 7 AL strategies have been included: Uniform, Diversity, Entropy, CoreSet, TypiClust, kVolIn and kVolOut (with $k=4$). The first three respectively belong to the density, the diversity and the uncertainty family. The next two (CoreSet, TypiClust) are strategies combining the assets of at least two AL families. The last two (kVolIn and kVolOut) are the new AL strategies (with $k=4$) respectively focusing on the input space for kVolIn, and the output space kVolOut.

\begin{figure}[h]
    \centering
    \begin{subfigure}[c]{0.8\linewidth}
    \centering
            \includegraphics[width=\linewidth]{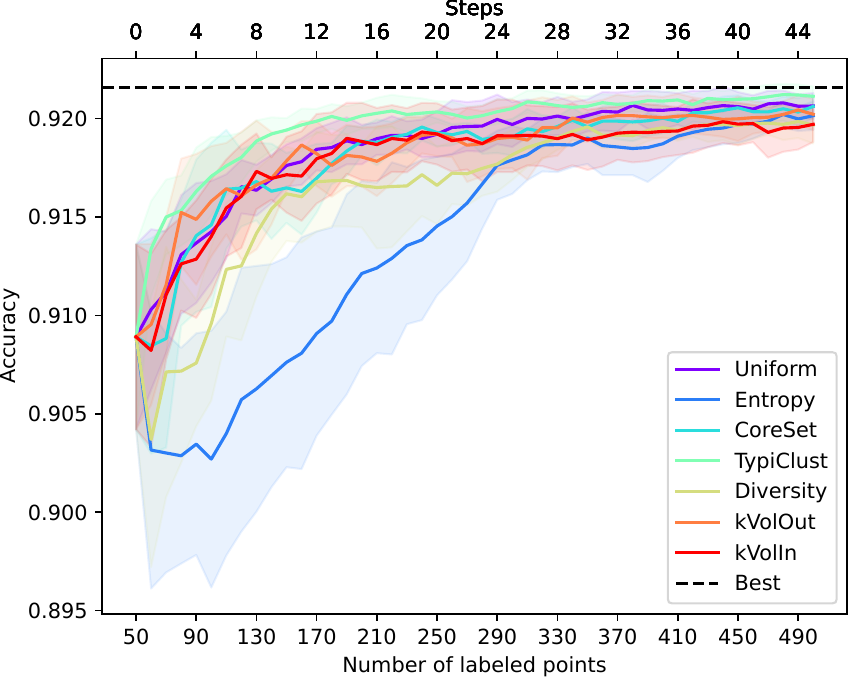}
            \caption{Scenario No. 1 (balanced)}
    \end{subfigure}
    \begin{subfigure}[c]{0.8\linewidth}
    \centering
            \includegraphics[width=\linewidth]{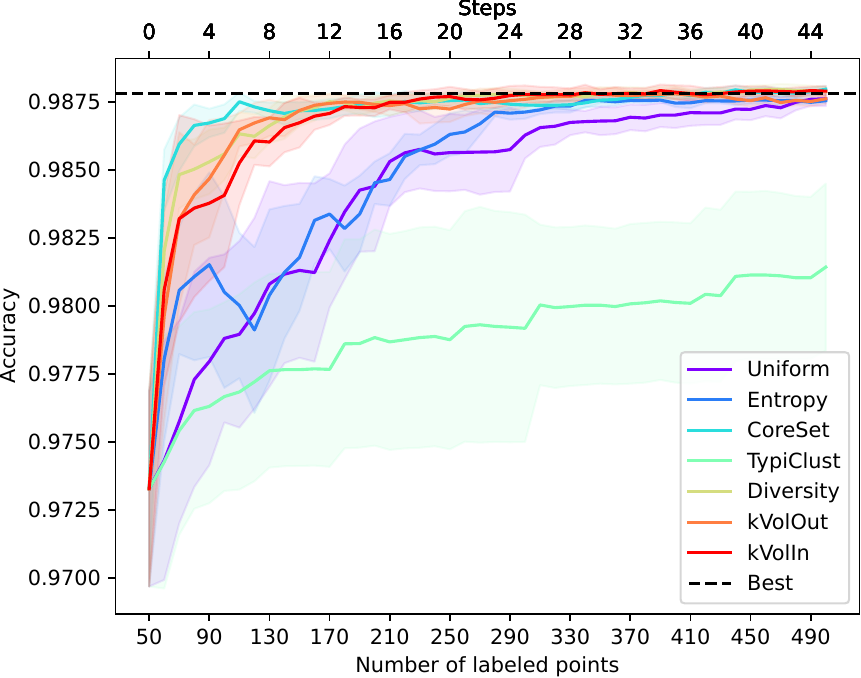}
            \caption{Scenario No. 2 (imbalanced)}
    \end{subfigure}
    \caption{Accuracy comparison during active training. Initial batch size $b_0=50$, step batch size $b=10$ and budget $B=500$. The dimension of the data space is 2.}
\label{fig:synthetic_exp_dim2_initbs50}
\end{figure}

\paragraph{Results.}
The accuracy is plotted against the cardinality of $\Dlab$ for $d=2$ on \cref{fig:synthetic_exp_dim2_initbs50}, and for $d=20$ on \cref{fig:synthetic_exp_dim20_initbs50}.
From \cref{fig:synthetic_exp_dim2_initbs50}, it appears that Uniform is more sensitive to the imbalance phenomenon. As an element of the density family, it is more influenced by any dominating class, which makes it perform worse that the others
(Scenario~No.2).
The situation seems similar for Entropy with curves close to the ones of Uniform in Scenarios~1 and~2.
The behavior of TypiClust is even more contrasted since it is the best AL strategy in Scenario~1, but completely fails with imbalanced classes since its curve is uniformly the lowest among the competitors (Scenario~2).
By contrast CoreSet, kVolIn, and kVolOut exhibit curves that are close to each other in these two scenarios.
The more difficult 20-dimensional input space scenario confirms these trends, as one can see from \cref{fig:synthetic_exp_dim20_initbs50} in \cref{app:synthetic_data}.
However two groups of AL strategies can be clearly identified now (Scenario~4): TypiClust (uniformly worst), Uniform, and kVolIn seem to behave poorly, whereas CoreSet, Entropy, Diversity and kVolOut perform equally well. (The same remark also holds in Scenario~3, but curves are less easy to distinguish from one another.)
This illustrates the effect of the high dimension of the input space. Compared to Scenario~2, the worse performance of kVolIn suggests it is more sensitive to the input dimension increase than kVolOut. Since the dimension of the output space remains unchanged compared to Scenarios~1 and~2, this justifies why the kVolOut performance does not vary after the input-space dimension has increased.

\subsection{Tabular data}
\label{sub:tabular_data}

\begin{table*}[h]
    \centering
    \caption{Characteristics of the tabular datasets used in \cref{sub:tabular_data}.}
    \label{tab:tab_exp_char}
    \begin{tabular}{lcccc}
    \toprule
        Dataset Id & Name & Number of classes & Number of points & Input dimension \\
    \midrule
         46      & splice      & 3 & 3190 & 60 \\
         182     & satimage    & 6 & 6430 & 36 \\
         846     & elevators   & 2 & 16599 & 18\\
    \bottomrule
    \end{tabular}
\end{table*}
The present section considers real-life tabular datasets taken from the openml project\footnote{\url{https://www.openml.org/}} through their Python API \citep{feurerOpenMLPythonExtensiblePython2021}. 
The selected datasets for this section are detailed in \cref{tab:tab_exp_char}.
Unlike \cref{sub:synth_data}, several classifiers are tested with the default parameter values from their Scikit-learn implementations \citep{pedregosaScikitlearnMachineLearning2011}:
Linear discriminant analysis (lda);
Multinomial logistic regression (mlr);
Support vector machine (svm);
Random forest (rf);
Multi layer perceptron (mlp). Let us notice that the compared AL strategies are now: Uniform, Entropy, CoreSet, TypiClust, kVolIn, and kVolOut.
Diversity has been excluded from the comparison since it exhibits similar performances to other already included strategies. 
The initial batch size is fixed to $b_0=10$ points and the step batch size $b=10$ as well. The maximum budget is $B=500$ samples.

\begin{figure}[h]
    \centering
    \begin{subfigure}[c]{0.8\linewidth}
    \centering
            \includegraphics[width=\linewidth]{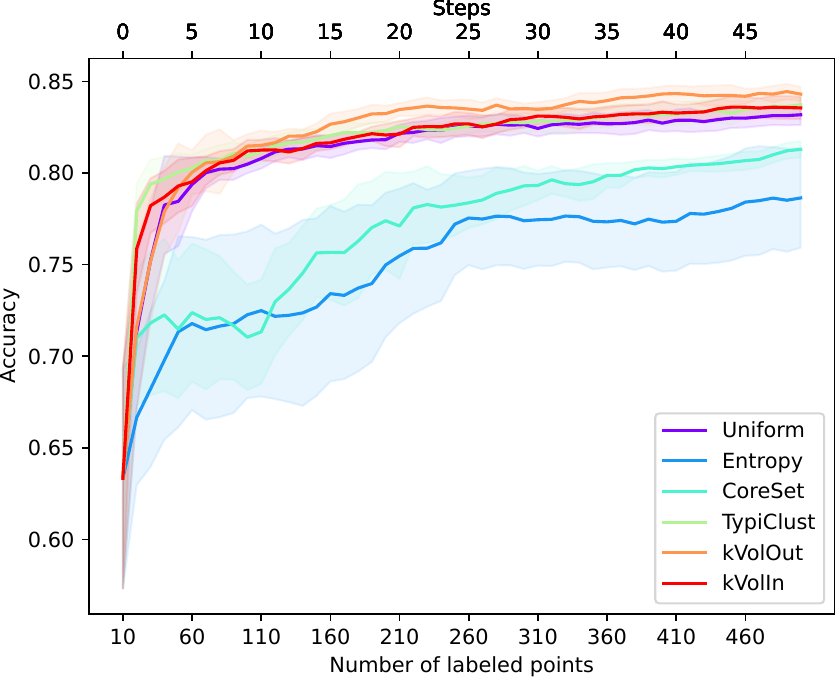}
            \caption{mlr.}
    \end{subfigure}
    \begin{subfigure}[c]{0.8\linewidth}
    \centering
            \includegraphics[width=\linewidth]{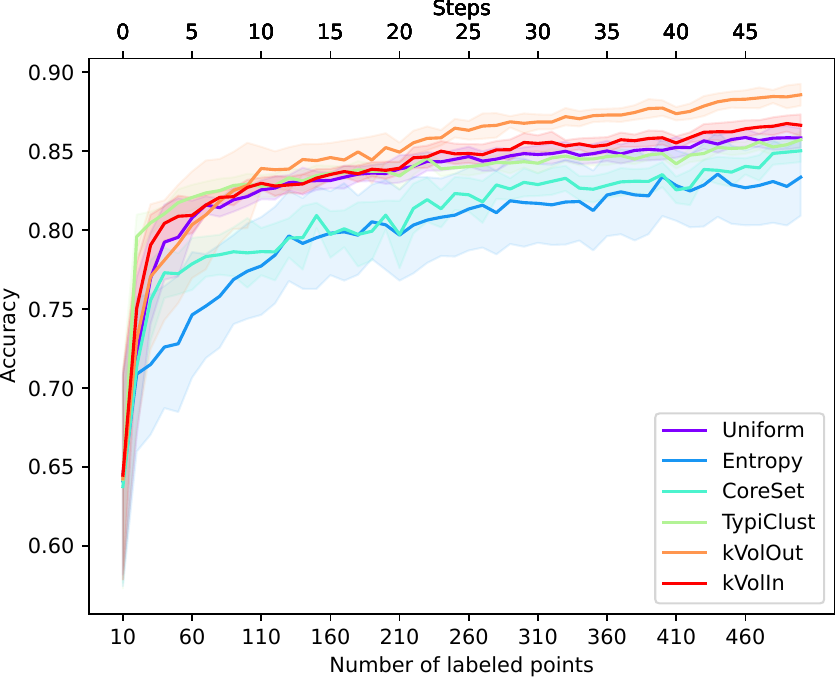}
            \caption{mlp.}
    \end{subfigure}
    \caption{
    Accuracy comparison during active training. Initial batch size $b_0=10$, step batch size $b=10$ and budget $B=500$.
    The dataset id is 182.}
    \label{fig:tab_data_id182_imbalanced_mlr_mlp}
\end{figure}

\Cref{fig:tab_data_id182_imbalanced_mlr_mlp} displays the results for dataset 182 (\cref{tab:tab_exp_char}) with the mlr and mlp classifiers.
The graphs illustrating the performance of the other classifiers have been deferred to \cref{app:tabular_data}.
%
Let us point out that a rebalanced version of each dataset has also been considered to check the results of \cref{sub:synth_data}.
Since the conclusions were unchanged, the results have not been included here.
%
\cref{fig:tab_data_id182_imbalanced_mlr_mlp} shows that the two classifiers (Panels~(a) and~(b)) exhibit the same behavior.
The worst performance is achieved by Entropy and CoreSet. The discrepancy with the other strategies is 
significant.
By contrast, Uniform, kVolIn, and TypiClust belong to a second group the accuracy of which reaches a high level. 
Eventually, it turns out that kVolOut achieves the top performance on this real-life dataset. For both classifiers, it is the best although the largest discrepancy is observed with mlp (Panel~(b)).

Combined with the results observed with the synthetic datasets from \cref{sub:synth_data}, all of this emphasizes the overall strong stability of the kVolOut strategy. It seems to lie among the AL strategies with best performance (accuracy) whatever the scenario or the underlying classifier, unlike other AL strategies whose performance varies a lot from one experiment to another (see CoreSet and TypiClust).
Furthermore, kVolOut strongly outperforms upon the other state-of-the-art competitors on real-life tabular datasets (see \cref{fig:tab_data_id182_imbalanced_mlr_mlp}, and \cref{fig:tab_data_id182_imbalanced_app} for instance).

\subsection{Image data}
\label{sub:image_data}

\begin{table*}[h]
    \centering
        \caption{Characteristics of the datasets of images used in \cref{sub:image_data}.}
    \label{tab:char_image_datasets}
    \begin{tabular}{lccccr}
    \toprule
          Name & Number of classes & Number of points & Image dimension & Embedding dimension & Reference\\
    \midrule
          MNIST      & 10 & 60000 & 784 & 2048 & \citep{lecun1998mnist}\\
          CIFAR10    & 10 & 50000 & 3072 & 2048 & \citep{krizhevskyLearningMultipleLayers2009}\\
          CIFAR100 & 100 & 50000 & 3072  & 2048 & \citep{krizhevskyLearningMultipleLayers2009}\\
    \bottomrule
    \end{tabular}
\end{table*}
The present section explores the performance of AL strategies on 3 classical Image datasets in high dimension (see \cref{tab:char_image_datasets} for a description). 
To get closer to real-life use-cases, the ResNet50\footnote{See \href{https://docs.pytorch.org/vision/main/models/generated/torchvision.models.resnet50.html\#torchvision.models.resnet50}{PyTorch documentation} and \citep{heDeepResidualLearning2015} more details.} model from PyTorch \citep{anselPyTorch2Faster2024} has been pre-trained on ImageNet \citep{dengImagenetLargescaleHierarchical2009} by loading the weights \texttt{ResNet50\_Weights.DEFAULT} and the associated \texttt{transforms}.
The last linear layer and the Softmax layer are discarded in order to use this model as an embedding network into a space of dimension 2048.
%
This allows to skip the training of the whole neural network from the ground up, which would complicate the evaluation of the active learning methods.

\begin{figure}[ht]
    \centering
    \begin{subfigure}[c]{0.8\linewidth}
    \centering
            \includegraphics[width=\linewidth]{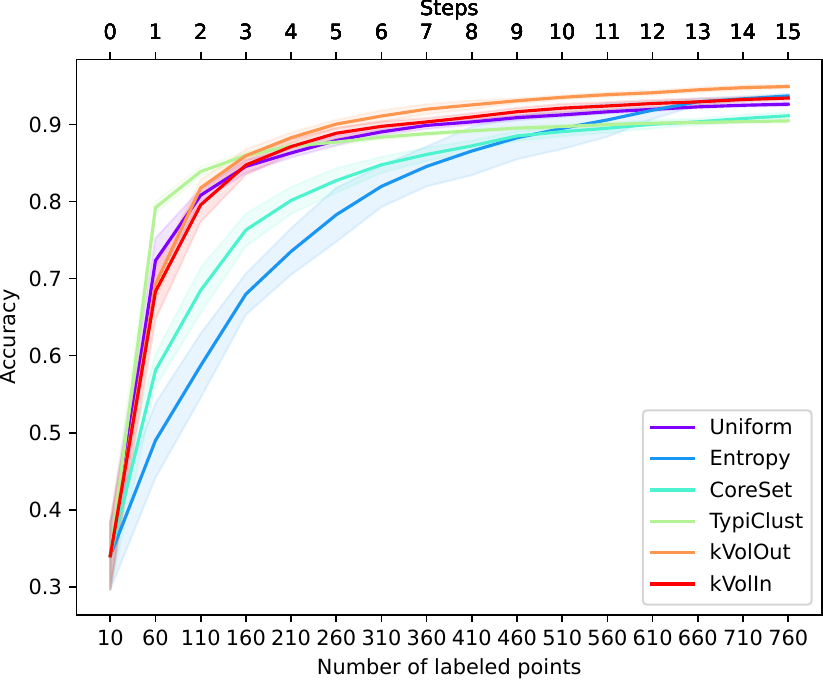}
            \caption{MNIST}
    \end{subfigure}
    \begin{subfigure}[c]{0.8\linewidth}
    \centering
            \includegraphics[width=\linewidth]{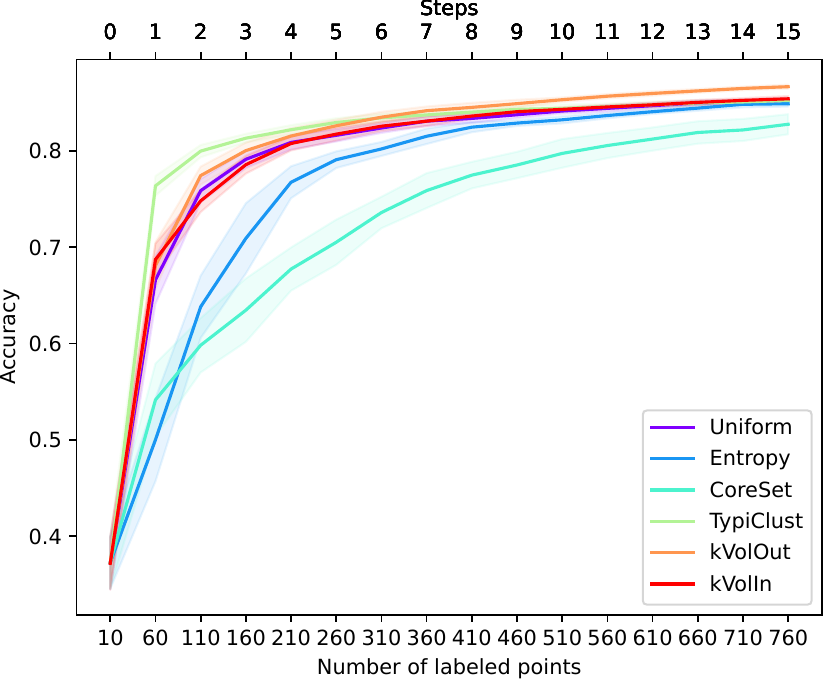}
            \caption{CIFAR10}
    \end{subfigure}
    \caption{Accuracy comparison during active training. Initial batch size $b_0=10$, step batch size $b=50$ and budget $B=760$.}
    \label{fig:image_data}
\end{figure}

\begin{figure}[h]
    \centering
    \centering
            \includegraphics[width=0.8\linewidth]{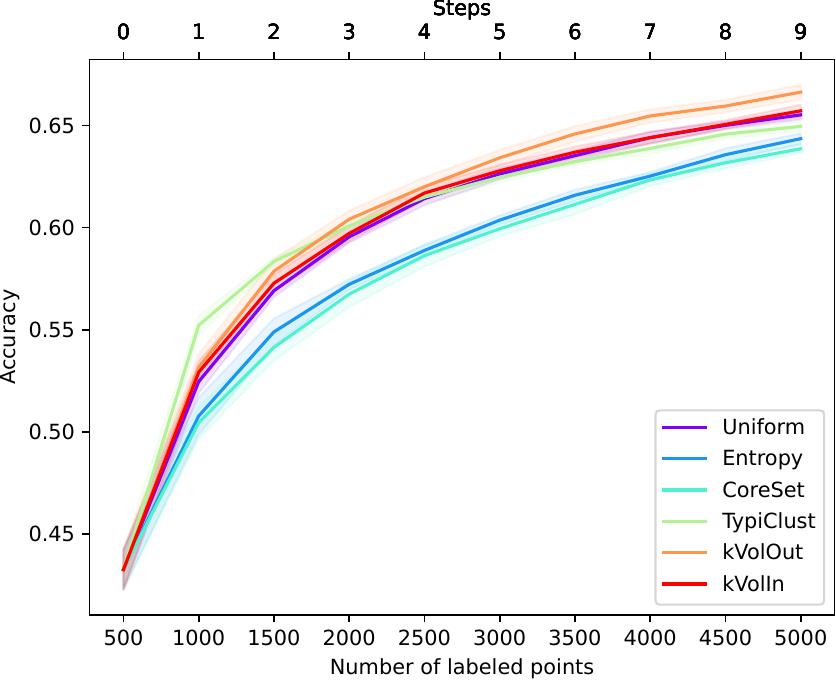}
    \caption{CIFAR100. Accuracy comparison during active training. Initial batch size $b_0=500$, step batch size $b=500$ and budget $B=5000$.}
    \label{fig:image_data_CIFAR100}
\end{figure}

\Cref{fig:image_data,fig:image_data_CIFAR100} display the performance of the same AL strategies as in \cref{sub:tabular_data} on MNIST (Panel~(a)) and CIFAR10 (Panel~(b)), and CIFAR100 (\cref{fig:image_data_CIFAR100}) datasets.
All panels of \cref{fig:image_data,fig:image_data_CIFAR100} confirm that the worst performances are jointly achieved by Entropy and CoreSet, although the Entropy performance seems to improve faster than the one of CoreSet.
All other methods seem to perform equivalently well, with nevertheless slight distinctions. Specifically, TypiClust reaches the best overall performance within a narrow interval which varies depending on the dataset. For instance it performs the best in the range $\abs{\Dlab} \in [60, 110]$ (MNIST).  
However as soon as one is allowed to go beyond this (somewhat narrow) range, then it turns out that kVolOut is the best AL strategy (see all panels of \cref{fig:image_data,fig:image_data_CIFAR100}).
\begin{remark}
    Although convergence is reached for MNIST and CIFAR10, one could worry that \cref{fig:image_data_CIFAR100} suggests it could not be the case for CIFAR100. 
    One could then argue that labeling 5\,000 points (10\% of the CIFAR100 dataset) is already highly time consuming and that the performances of AL strategies at this point are already representative of a realistic situation.
    Nevertheless experiments with a maximum of 10\,000 points (see \cref{fig:image_data_CIFAR100_10000} in \cref{app:image_data}) have been added for the sake of completeness. 
    Fortunately, our previous conclusion that kVolOut is the best AL strategy on this dataset remains unchanged.
    However for computational reasons, we had to remove CoreSet and TypiClust from the comparison since they were highly time consuming compared to kVolOut.
\end{remark}

\paragraph{Computing time.}
One major asset of the kVol strategy is its computational efficiency.
The kVolOut computing time corresponding to the experiment reported in \cref{fig:image_data_CIFAR100} is of the same order as the one of the Uniform and Entropy strategies.
Even more remarkably, kVolOut is also 1.8 (resp. 6.2) times faster than CoreSet (resp. TypiClust). 
The precise computing times are summarized by \cref{tab:computing_time_comparison} in \cref{app:computing_time_comparison}.
At step $t\in\bra{T}$, the computation of $\alpha_{\kvol}$ can be divided in two phases: $(i)$ Computing $\kvol$ for each point in $\Dpool_{t-1}$; $(ii)$ Sampling $b$ indices from a categorical distribution.
Phase $(i)$ can use brute force in $O\pa{dkn^2}$ time due to the pairwise distance matrix and a negligible number of multiplications. This can be reduced when dimension $d$ is small \citep{vaidyaAnOnLognAlgorithm1989}, which is especially relevant for kVolOut.
Phase $(ii)$ can then be performed with the \emph{alias method} \citep{voseLinearAlgorithmGenerating1991} in $O(n)$ time.
The total time of a $k$-Volume query is thus in $O(dkn^2)$.


\clearpage
\subsection*{AI use statement}


In this work, we used generative AI tools for [assist with translation, assist in the writing of proofs].
We have not used generative AI tools for [Generate synthetic data sets, help develop theoretical models or conceptual frameworks, formulate mathematical claims, provide critical ingredients for proving mathematical claims, propose or refine hypotheses, design or provide feedback on research  methodology or experiments, implement methods, clean and reformat dataset, support qualitative and thematic data analysis, interpret results]. 
Additionally, we used generative AI tools for [identify some relevant literature].
We have reviewed all AI-assisted work.
We take responsibility for the final content of this work, including text, claims, or artifacts produced with the aid of generative AI.




\subsection*{Acknowledgements}
This work benefited from financial support from the French government managed by the National Agency for Research under the France 2030 program, with the reference ANR-23-PEIA-0004.

\printbibliography

@article{citovsky2021batch,
  title={Batch active learning at scale},
  author={Citovsky, Gui and DeSalvo, Giulia and Gentile, Claudio and Karydas, Lazaros and Rajagopalan, Anand and Rostamizadeh, Afshin and Kumar, Sanjiv},
  journal={Advances in Neural Information Processing Systems},
  volume={34},
  pages={11933--11944},
  year={2021}
}

@misc{liSurveyDeepActive2024,
  title = {A {{Survey}} on {{Deep Active Learning}}: {{Recent Advances}} and {{New Frontiers}}},
  shorttitle = {A {{Survey}} on {{Deep Active Learning}}},
  author = {Li, Dongyuan and Wang, Zhen and Chen, Yankai and Jiang, Renhe and Ding, Weiping and Okumura, Manabu},
  year = 2024,
  month = jul,
  number = {arXiv:2405.00334},
  eprint = {2405.00334},
  primaryclass = {cs},
  publisher = {arXiv},
  doi = {10.48550/arXiv.2405.00334},
  archiveprefix = {arXiv},
  langid = {english}
}

@article{wangDeepActiveLearning2022,
  title = {Deep {{Active Learning}} by {{Leveraging Training Dynamics}}},
  author = {Wang, Haonan and Huang, Wei and Wu, Ziwei and Margenot, Andrew and Tong, Hanghang and He, Jingrui},
  year = 2022,
  journal = {Proceedings of the 36th International Conference on Neural Information Processing Systems},
  series = {{{NIPS}} '22},
  issn = {9781713871088},
  doi = {10.5555/3600270.3602095},
  langid = {english}
}

@misc{popDeepEnsembleBayesian2018,
  title = {Deep {{Ensemble Bayesian Active Learning}} : {{Addressing}} the {{Mode Collapse}} Issue in {{Monte Carlo}} Dropout via {{Ensembles}}},
  shorttitle = {Deep {{Ensemble Bayesian Active Learning}}},
  author = {Pop, Remus and Fulop, Patric},
  year = 2018,
  month = nov,
  number = {arXiv:1811.03897},
  eprint = {1811.03897},
  primaryclass = {cs},
  publisher = {arXiv},
  doi = {10.48550/arXiv.1811.03897},
  archiveprefix = {arXiv}
}

@article{kirschUnifyingApproachesActive2022,
  title = {Unifying {{Approaches}} in {{Active Learning}} and {{Active Sampling}} via {{Fisher Information}} and {{Information-Theoretic Quantities}}},
  author = {Kirsch, Andreas and Gal, Yarin},
  year = 2022,
  month = aug,
  journal = {Transactions on Machine Learning Research},
  issn = {2835-8856},
  doi = {10.48550/arxiv.2208.00549},
  langid = {english}
}

@inproceedings{gulFisherMaskEnhancingNeural2024,
  title = {{{FisherMask}}: {{Enhancing Neural Network Labeling Efficiency}} in {{Image Classification Using Fisher Information}}},
  shorttitle = {{{FisherMask}}},
  booktitle = {2024 {{IEEE International Conference}} on {{Big Data}} ({{BigData}})},
  author = {Gul, Shreen and Elmahallawy, Mohamed and Madria, Sanjay and Tripathy, Ardhendu},
  year = 2024,
  month = dec,
  pages = {1302--1308},
  issn = {2573-2978},
  doi = {10.1109/BigData62323.2024.10825241}
}

@inproceedings{senerActiveLearningConvolutional2018,
  title = {Active {{Learning}} for {{Convolutional Neural Networks}}: {{A Core-Set Approach}}},
  shorttitle = {Active {{Learning}} for {{Convolutional Neural Networks}}},
  booktitle = {International {{Conference}} on {{Learning Representations}}},
  author = {Sener, Ozan and Savarese, Silvio},
  year = 2018,
  month = feb,
  langid = {english}
}

@inproceedings{munjalRobustReproducibleActive2022,
  title = {Towards {{Robust}} and {{Reproducible Active Learning}} Using {{Neural Networks}}},
  booktitle = {2022 {{IEEE}}/{{CVF Conference}} on {{Computer Vision}} and {{Pattern Recognition}} ({{CVPR}})},
  author = {Munjal, Prateek and Hayat, Nasir and Hayat, Munawar and Sourati, Jamshid and Khan, Shadab},
  year = 2022,
  month = jun,
  pages = {223--232},
  issn = {2575-7075},
  doi = {10.1109/CVPR52688.2022.00032}
}

@article{elreedyNovelActiveLearning2019,
  title = {A {{Novel Active Learning Regression Framework}} for {{Balancing}} the {{Exploration-Exploitation Trade-Off}}},
  author = {Elreedy, Dina and Atiya, Amir F. and Shaheen, Samir I.},
  year = 2019,
  month = jul,
  journal = {Entropy},
  volume = {21},
  number = {7},
  publisher = {publisher},
  issn = {1099-4300},
  doi = {10.3390/e21070651},
  copyright = {http://creativecommons.org/licenses/by/3.0/},
  langid = {english}
}

@inproceedings{krauseNonmyopicActiveLearning2007,
  title = {Nonmyopic Active Learning of {{Gaussian}} Processes: An Exploration-Exploitation Approach},
  shorttitle = {Nonmyopic Active Learning of {{Gaussian}} Processes},
  booktitle = {Proceedings of the 24th International Conference on {{Machine}} Learning},
  author = {Krause, Andreas and Guestrin, Carlos},
  year = 2007,
  month = jun,
  series = {{{ICML}} '07},
  pages = {449--456},
  publisher = {Association for Computing Machinery},
  address = {New York, NY, USA},
  doi = {10.1145/1273496.1273553},
  isbn = {978-1-59593-793-3}
}

@unpublished{bouneffoufContextualBanditActive2014,
  title = {Contextual {{Bandit}} for {{Active Learning}}: {{Active Thompson Sampling}}},
  shorttitle = {Contextual {{Bandit}} for {{Active Learning}}},
  author = {Bouneffouf, Djallel and Laroche, Romain and Urvoy, Tanguy and F{\'e}raud, Raphael and Allesiardo, Robin},
  year = 2014,
  month = sep
}

@article{bouneffoufExponentiatedGradientExploration2016,
  title = {Exponentiated {{Gradient Exploration}} for {{Active Learning}}},
  author = {Bouneffouf, Djallel},
  year = 2016,
  month = jan,
  journal = {Computers},
  volume = {5},
  number = {1},
  publisher = {publisher},
  issn = {2073-431X},
  doi = {10.3390/computers5010001},
  copyright = {http://creativecommons.org/licenses/by/3.0/},
  langid = {english}
}

@article{keeQuerybycommitteeImprovementDiversity2018,
  title = {Query-by-Committee Improvement with Diversity and Density in Batch Active Learning},
  author = {Kee, Seho and {del Castillo}, Enrique and Runger, George},
  year = 2018,
  month = jul,
  journal = {Information Sciences},
  volume = {454--455},
  pages = {401--418},
  issn = {0020-0255},
  doi = {10.1016/j.ins.2018.05.014}
}

@misc{boucheron2013concentration,
  title={Concentration inequalities: A nonasymptotic theory of independence oxford, uk: Oxford univ},
  author={Boucheron, S and Lugosi, G and Massart, P},
  year={2013},
  publisher={Press}
}

@misc{eisenhardtMindGapFramework2026,
  title = {Mind the {{Gap}}: {{A Framework}} for {{Assessing Pitfalls}} in {{Multimodal Active Learning}}},
  shorttitle = {Mind the {{Gap}}},
  author = {Eisenhardt, Dustin and Jeong, Yunhee and Buettner, Florian},
  year = 2026,
  publisher = {arXiv},
  doi = {10.48550/ARXIV.2603.29677},
  copyright = {Creative Commons Attribution Share Alike 4.0 International}
}

@article{cebronActiveLearningObject2009,
  title = {Active Learning for Object Classification: From Exploration to Exploitation},
  shorttitle = {Active Learning for Object Classification},
  author = {Cebron, Nicolas and Berthold, Michael R.},
  year = 2009,
  month = apr,
  journal = {Data Mining and Knowledge Discovery},
  volume = {18},
  number = {2},
  pages = {283--299},
  issn = {1573-756X},
  doi = {10.1007/s10618-008-0115-0},
  langid = {english}
}

@inproceedings{huExploringLimitationsActive2021,
  title = {Towards {{Exploring}} the {{Limitations}} of {{Active Learning}}: {{An Empirical Study}}},
  shorttitle = {Towards {{Exploring}} the {{Limitations}} of {{Active Learning}}},
  booktitle = {2021 36th {{IEEE}}/{{ACM International Conference}} on {{Automated Software Engineering}} ({{ASE}})},
  author = {Hu, Qiang and Guo, Yuejun and Cordy, Maxime and Xie, Xiaofei and Ma, Wei and Papadakis, Mike and Traon, Yves Le},
  year = 2021,
  month = nov,
  pages = {917--929},
  issn = {2643-1572},
  doi = {10.1109/ASE51524.2021.9678672}
}

@techreport{settlesActiveLearningLiterature2009,
  type = {Technical {{Report}}},
  title = {Active {{Learning Literature Survey}}},
  author = {Settles, Burr},
  year = 2009,
  institution = {University of Wisconsin-Madison Department of Computer Sciences},
  langid = {english}
}

@article{wangCostEffectiveActiveLearning2017,
  title = {Cost-{{Effective Active Learning}} for {{Deep Image Classification}}},
  author = {Wang, Keze and Zhang, Dongyu and Li, Ya and Zhang, Ruimao and Lin, Liang},
  year = 2017,
  month = dec,
  journal = {IEEE Transactions on Circuits and Systems for Video Technology},
  volume = {27},
  number = {12},
  eprint = {1701.03551},
  primaryclass = {cs},
  pages = {2591--2600},
  issn = {1051-8215, 1558-2205},
  doi = {10.1109/TCSVT.2016.2589879},
  archiveprefix = {arXiv}
}

@inproceedings{xuIncorporatingDiversityDensity2007,
  title = {Incorporating {{Diversity}} and {{Density}} in {{Active Learning}} for {{Relevance Feedback}}},
  booktitle = {Advances in {{Information Retrieval}}},
  author = {Xu, Zuobing and Akella, Ram and Zhang, Yi},
  editor = {Amati, Giambattista and Carpineto, Claudio and Romano, Giovanni},
  year = 2007,
  pages = {246--257},
  publisher = {Springer},
  address = {Berlin, Heidelberg},
  doi = {10.1007/978-3-540-71496-5_24},
  isbn = {978-3-540-71496-5},
  langid = {english}
}

@article{gaoComprehensiveSurveyImbalanced2026,
  title = {A {{Comprehensive Survey}} on {{Imbalanced Data Learning}}},
  author = {Gao, Xinyi and Xie, Dongting and Zhang, Yihang and Wang, Zhengren and Chen, Chong and He, Conghui and Yin, Hongzhi and Zhang, Wentao},
  year = 2026,
  month = nov,
  journal = {Frontiers of Computer Science},
  volume = {20},
  number = {11},
  eprint = {2502.08960},
  primaryclass = {cs.LG},
  pages = {2011622},
  issn = {2095-2228, 2095-2236},
  doi = {10.1007/s11704-025-50274-7},
  archiveprefix = {arXiv},
  langid = {english}
}

@article{chenSurveyImbalancedLearning2024,
  title = {A Survey on Imbalanced Learning: Latest Research, Applications and Future Directions},
  shorttitle = {A Survey on Imbalanced Learning},
  author = {Chen, Wuxing and Yang, Kaixiang and Yu, Zhiwen and Shi, Yifan and Chen, C. L. Philip},
  year = 2024,
  month = may,
  journal = {Artificial Intelligence Review},
  volume = {57},
  number = {6},
  pages = {137},
  issn = {1573-7462},
  doi = {10.1007/s10462-024-10759-6},
  langid = {english}
}

@inproceedings{hacohenActiveLearningBudget2022,
  title = {Active {{Learning}} on a {{Budget}}: {{Opposite Strategies Suit High}} and {{Low Budgets}}},
  shorttitle = {Active {{Learning}} on a {{Budget}}},
  booktitle = {Proceedings of the 39th {{International Conference}} on {{Machine Learning}}},
  author = {Hacohen, Guy and Dekel, Avihu and Weinshall, Daphna},
  year = 2022,
  month = jun,
  pages = {8175--8195},
  publisher = {PMLR},
  issn = {2640-3498},
  langid = {english}
}

@book{biauLecturesNearestNeighbor2015,
  title = {Lectures on the {{Nearest Neighbor Method}}},
  author = {Biau, G{\'e}rard and Devroye, Luc},
  year = 2015,
  series = {Springer {{Series}} in the {{Data Sciences}}},
  publisher = {Springer International Publishing},
  address = {Cham},
  doi = {10.1007/978-3-319-25388-6},
  copyright = {http://www.springer.com/tdm},
  langid = {english}
}

@article{krizhevskyLearningMultipleLayers2009,
  title = {Learning Multiple Layers of Features from Tiny Images},
  author = {Krizhevsky, Alex and Hinton, Geoffrey},
  year = 2009,
  publisher = {Toronto, ON, Canada}
}

@inproceedings{anselPyTorch2Faster2024,
  title = {{{PyTorch}} 2: {{Faster Machine Learning Through Dynamic Python Bytecode Transformation}} and {{Graph Compilation}}},
  shorttitle = {{{PyTorch}} 2},
  booktitle = {Proceedings of the 29th {{ACM International Conference}} on {{Architectural Support}} for {{Programming Languages}} and {{Operating Systems}}, {{Volume}} 2},
  author = {Ansel, Jason and Yang, Edward and He, Horace and Gimelshein, Natalia and Jain, Animesh and Voznesensky, Michael and Bao, Bin and Bell, Peter and Berard, David and Burovski, Evgeni and Chauhan, Geeta and Chourdia, Anjali and Constable, Will and Desmaison, Alban and DeVito, Zachary and Ellison, Elias and Feng, Will and Gong, Jiong and Gschwind, Michael and Hirsh, Brian and Huang, Sherlock and Kalambarkar, Kshiteej and Kirsch, Laurent and Lazos, Michael and Lezcano, Mario and Liang, Yanbo and Liang, Jason and Lu, Yinghai and Luk, C. K. and Maher, Bert and Pan, Yunjie and Puhrsch, Christian and Reso, Matthias and Saroufim, Mark and Siraichi, Marcos Yukio and Suk, Helen and Zhang, Shunting and Suo, Michael and Tillet, Phil and Zhao, Xu and Wang, Eikan and Zhou, Keren and Zou, Richard and Wang, Xiaodong and Mathews, Ajit and Wen, William and Chanan, Gregory and Wu, Peng and Chintala, Soumith},
  year = 2024,
  month = apr,
  series = {{{ASPLOS}} '24},
  volume = {2},
  pages = {929--947},
  publisher = {Association for Computing Machinery},
  address = {New York, NY, USA},
  doi = {10.1145/3620665.3640366},
  isbn = {979-8-4007-0385-0}
}

@misc{lecun1998mnist,
  title = {The {{MNIST}} Database of Handwritten Digits},
  author = {LeCun, Yann},
  year = 1998
}

@article{feurerOpenMLPythonExtensiblePython2021,
  title = {{{OpenML-Python}}: An Extensible {{Python API}} for {{OpenML}}},
  shorttitle = {{{OpenML-Python}}},
  author = {Feurer, Matthias and van Rijn, Jan N. and Kadra, Arlind and Gijsbers, Pieter and Mallik, Neeratyoy and Ravi, Sahithya and M{\"u}ller, Andreas and Vanschoren, Joaquin and Hutter, Frank},
  year = 2021,
  journal = {Journal of Machine Learning Research},
  volume = {22},
  number = {100},
  pages = {1--5},
  issn = {1533-7928}
}

@article{pedregosaScikitlearnMachineLearning2011,
  title = {Scikit-Learn: {{Machine Learning}} in {{Python}}},
  shorttitle = {Scikit-Learn},
  author = {Pedregosa, Fabian and Varoquaux, Ga{\"e}l and Gramfort, Alexandre and Michel, Vincent and Thirion, Bertrand and Grisel, Olivier and Blondel, Mathieu and Prettenhofer, Peter and Weiss, Ron and Dubourg, Vincent and Vanderplas, Jake and Passos, Alexandre and Cournapeau, David and Brucher, Matthieu and Perrot, Matthieu and Duchesnay, {\'E}douard},
  year = 2011,
  journal = {Journal of Machine Learning Research},
  volume = {12},
  number = {85},
  pages = {2825--2830},
  issn = {1533-7928}
}

@inproceedings{dengImagenetLargescaleHierarchical2009,
  title = {Imagenet: {{A}} Large-Scale Hierarchical Image Database},
  booktitle = {2009 {{IEEE}} Conference on Computer Vision and Pattern Recognition},
  author = {Deng, Jia and Dong, Wei and Socher, Richard and Li, Li-Jia and Li, Kai and {Fei-Fei}, Li},
  year = 2009,
  pages = {248--255},
  publisher = {IEEE}
}

@book{casilliAttendantRobotsEnquete2019,
  title = {En attendant les robots. Enquête sur le travail du clic},
  author = {Casilli, Antonio},
  date = {2019},
  publisher = {Le Seuil},
  url = {https://shs.cairn.info/en-attendant-les-robots-enquete-sur-le-travail-du-clic--9782021401882},
  isbn = {978-2-02-140188-2},
  langid = {french}
}

@online{heDeepResidualLearning2015,
  title = {Deep {{Residual Learning}} for {{Image Recognition}}},
  author = {He, Kaiming and Zhang, Xiangyu and Ren, Shaoqing and Sun, Jian},
  date = {2015-12-10},
  eprint = {1512.03385},
  eprinttype = {arXiv},
  eprintclass = {cs.CV},
  doi = {10.48550/arXiv.1512.03385},
  pubstate = {prepublished}
}

@article{voseLinearAlgorithmGenerating1991,
  title = {A Linear Algorithm for Generating Random Numbers with a given Distribution},
  author = {Vose, M.D.},
  date = {1991-09},
  journaltitle = {IEEE Transactions on Software Engineering},
  shortjournal = {IEEE Trans. Softw. Eng.},
  volume = {17},
  number = {9},
  pages = {972--975},
  issn = {1939-3520},
  doi = {10.1109/32.92917}
}

@article{vaidyaAnOnLognAlgorithm1989,
  title = {{{AnO}}(n Logn) Algorithm for the All-Nearest-Neighbors {{Problem}}},
  author = {Vaidya, Pravin M.},
  date = {1989-03-01},
  journaltitle = {Discrete \& Computational Geometry},
  shortjournal = {Discrete Comput. Geom.},
  volume = {4},
  number = {2},
  pages = {101--115},
  issn = {1432-0444},
  doi = {10.1007/BF02187718},
  langid = {english}
}

@book{gyorfiDistributionFreeTheoryNonparametric2002,
  title = {A {{Distribution-Free Theory}} of {{Nonparametric Regression}}},
  author = {Györfi, László and Kohler, Michael and Krzyżak, Adam and Walk, Harro},
  date = {2002},
  series = {Springer {{Series}} in {{Statistics}}},
  publisher = {Springer},
  location = {New York, NY},
  doi = {10.1007/b97848},
  isbn = {978-0-387-95441-7 978-0-387-22442-8}
}

\clearpage
\appendix
\thispagestyle{empty}

\onecolumn

\section{Appendix}

\subsection{Shortcoming of Uncertainty Active Learning}
\label{app:shortcoming_uncertainty_AL}

\begin{figure}[h]
   \centering
   \includegraphics[width=0.6\linewidth]{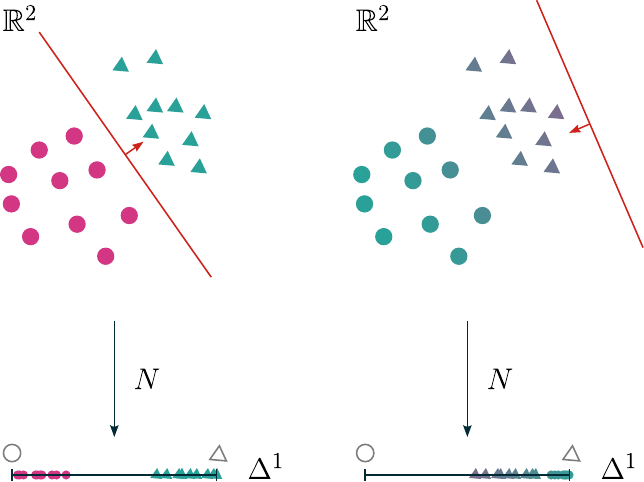}
   \caption{Difficult case for the uncertainty family. The red line represents the decision boundary of the trained classifier $N$, the colors represent the probabilities of each class, and the shape of the points represent the true labels.}
   \label{fig:counter_example_proba_based_al}
\end{figure}

The shortcoming mentioned in the introduction of the Uncertainty methods can be seen on \cref{fig:counter_example_proba_based_al}. The left-hand side shows a configuration for which uncertainty methods are beneficial, whereas the right-hand side of the figure shows that a wrongly positioned decision boundary may result in the selection of only triangles as they are the less certain predictions, even if they are the only correct ones when all the labels are available. 

\subsection{Proof of \cref{thm:conv_to_uniform}}
\label{app:proof_of_conv_to_uniform}

\begin{proof}
Let $\cD_n = \set{Z_1,...,Z_n}$ be the collection of $n$ i.i.d. $d$-dimensional random vectors with probability distribution $\mu$ with density $\rho$ as defined in \cref{thm:conv_to_uniform}. 
Let $\hat{\rho}_{k,n}$ be the $k$-NN-based density estimator of $\rho$ (see \citet{biauLecturesNearestNeighbor2015}) defined by
\[
				\hat{\rho}_{k,n}(\kvv) = \frac{k}{n} \frac{1}{\kvol_k(\kvv,\cD_n)} \\
.\]

According to \cref{def.KVol.query} with $b=1$, we randomly sample an index $I_{k,n}\in \set{1,...,n}$ with respect to the following probability distribution
\[
				\P\pa{I_{k,n}=i\mid \KVV_1,\ldots,\KVV_n} = \frac{{\hat{\rho}_{k,n}(\KVV_i)}^{-1}}{\sum_{j=1}^n {\hat{\rho}_{k,n}(\KVV_j)}^{-1}}
.\] 
In the following, we omit the subscript $k,n$ on $I$ or $\hat{\rho}$ to lighten the calculations.

Let $A$ be a Borel set included in $S := \Supp \mu$. Then, 
\begin{align*}
\P\pa{\KVV_I \in A} &= \E\bra{\1_{\set{\KVV_I\in A}}} \\
                 &= \E\bra{\E\bra{\1_{\set{\KVV_I\in A}} \mid \KVV_1,\ldots,\KVV_n}} \\
                 &= \E\bra{\sum_{i=1}^n\E\bra{\1_{\set{\KVV_i\in A}}\1_{\set{I = i}} \mid \KVV_1,\ldots,\KVV_n}} .
\end{align*}                                                 
Since $ \1_{\set{\KVV_i \in A}}$ is $\sigma$-measurable with respect to $\sigma\pa{\KVV_1,\ldots,\KVV_n}$, we have
\begin{align*}
\P\pa{\KVV_I \in A}				 &= \E\bra{\sum_{i=1}^n\1_{\set{\KVV_i\in A}}\E\bra{\1_{\set{I = i}} \mid \KVV_1,\ldots,\KVV_n}} \\
     &= \E\bra{\sum_{i=1}^n\1_{\set{\KVV_i\in A}}\P\pa{I = i \mid \KVV_1,\ldots,\KVV_n}} \\
     &= \E\bra{\frac{\sum_{i=1}^n\1_{\set{\KVV_i\in A}}{\hat{\rho}(\KVV_i)}^{-1}}{\sum_{j=1}^n {\hat{\rho}(\KVV_j)}^{-1}}} .
 \end{align*}

Let us introduce $\hat{N}_{k,n} = \frac{1}{n} \sum_{i=1}^n\1_{\set{\KVV_i\in A}}{\hat{\rho}_{k,n}(\KVV_i)}^{-1}$ and $\hat{D}_{k,n} = \frac{1}{n}\sum_{j=1}^n {\hat{\rho}_{k,n}(\KVV_j)}^{-1}$, and let $N_n$ and $D_n$ be their counterparts where $\hat{\rho}$ is replaced by $\rho$.
In the following, the focus is given to the numerator since the denominator has a similar behavior.

Let us first notice that the strong law of large numbers yields 
\begin{align}\label{eq.almost.sure.CV.Nn}
    N_n \xra[a.s.]{n\to\infty} \E\bra{\1_{\set{\KVV_1\in A}}{\rho(\KVV_1)}^{-1}} =  \int_A {{\rho(\kvv)}^{-1}} {\rho(\kvv)} d\kvv = \Vol(A) .
\end{align}
It remains to prove that the convergence still holds true with $\hat{N}_{k,n}$.

The idea of the remaining part of the proof consists in first introducing the set $L_m = \acc{z\in S \mid \rho(z) \leq m}$, for some $m>0$, and then splitting $A$ into two disjoint subsets, that is 
\begin{align*}
    A = \pa{A\cap \pa{L_m}^c} \cup \pa{A\cap L_m}.
\end{align*}
It results that
\begin{align*}
    \hat{N}_{k,n}  & = \frac{1}{n} \sum_{i=1}^n\1_{\set{\KVV_i\in A}}{\hat{\rho}_{k,n}(\KVV_i)}^{-1} \\
    & = \frac{1}{n} \sum_{i=1}^n\1_{\set{\KVV_i\in A\cap (L_m)^c}}{\hat{\rho}_{k,n}(\KVV_i)}^{-1} + \frac{1}{n} \sum_{i=1}^n\1_{\set{\KVV_i\in A \cap L_m}}{\hat{\rho}_{k,n}(\KVV_i)}^{-1} \\
    & = \hat{N}_{k,n}^1 + \hat{N}_{k,n}^2.
\end{align*}

\subsubsection*{Regions of high density.}
Let us consider $\hat{N}_{k,n}^1$ and $N_n^1 =\frac{1}{n} \sum_{i=1}^n\1_{\set{\KVV_i\in A\cap (L_m)^c}}{\rho(\KVV_i)}^{-1} $.
Simple calculations yield
\begin{align*}
    \hat{N}_{k,n}^1  & = \frac{1}{n} \sum_{i=1}^n \1_{\set{\KVV_i\in A\cap \pa{L_m}^c}} \pa{\frac{\rho(\KVV_i) - {\hat{\rho}(\KVV_i)}}{\rho(\KVV_i)\hat{\rho}(\KVV_i)}} + N_n^1 \\
     &\leq \underbrace{\frac{1}{n} \sum_{i=1}^n \1_{\set{\KVV_i\in A\cap \pa{L_m}^c}} {\hat{\rho}(\KVV_i)}^{-1}}_{\hat{N}_{k,n}^1} \rho(\KVV_i)^{-1} \max_{1\leq j\leq n} \abs{\rho(\KVV_j) - {\hat{\rho}(\KVV_j)}}  + N_n^1  \\
     &\leq \hat{N}_{k,n}^1 \frac{ \max_{1\leq j\leq n} \abs{\rho(\KVV_j) - {\hat{\rho}(\KVV_j)}} }{m}   + N_n^1  ,
\end{align*}
which implies that
\begin{align*}
\hat{N}_{k,n}^1\pa{1 - \frac{ \max_{1\leq j\leq n} \abs{\rho(\KVV_j) - {\hat{\rho}(\KVV_j)}} }{m} } \leq  N_n^1 .
\end{align*}    
The same reasoning also leads to
\begin{align*}
N_n^1 \leq 	\hat{N}_{k,n}^1\pa{1 + \frac{ \max_{1\leq j\leq n} \abs{\rho(\KVV_j) - {\hat{\rho}(\KVV_j)}} }{m}  } .    
\end{align*}

Further recalling that $\rho$ is uniformly continuous, $\frac{k}{\log n} \to \infty$, and $\frac{k}{n} \to 0$, then \citet[Theorem 4.2]{biauLecturesNearestNeighbor2015} implies that
\[
    \sup_{\kvv\in\R^d} \abs{\rho(\kvv) - \hat{\rho}(\kvv)} \xra[a.s.]{n\to\infty} 0.
\] 
Now assuming $n$ large enough such that $\sup_\kvv \abs{\rho(\kvv)-\hat{\rho}(\kvv)} \frac{1}{m} < 1$ almost surely, it results that
\begin{align} \label{eq.sandwich.inequality}
    \frac{N_n^1}{1 + \sup_\kvv \abs{\rho(\kvv)-\hat{\rho}(\kvv)} \frac{1}{m}} \leq \hat{N}_{k,n}^1 \leq \frac{N_n^1}{1 - \sup_\kvv \abs{\rho(\kvv)-\hat{\rho}(\kvv)} \frac{1}{m}} .
\end{align}

By gathering all previous results, both denominators converge to $1$ almost surely, and the law of large numbers yields that 
\begin{align}\label{eq.N1.almost.sure}
    N_n^1\to \Vol(A\cap (L_m)^c)
\end{align} almost surely as $n\to +\infty$, then leading to the same limit for $\hat{N}_{k,n}^1$. 

\subsubsection*{Regions of low density.}
The quantity $\hat{N}_{k,n}^2$ can be rewritten as
\begin{align*}
    \hat{N}_{k,n}^2  &=  \frac{1}{n} \sum_{i=1}^n\1_{\set{\KVV_i\in A\cap L_m}}{\hat{\rho}_{k,n}(\KVV_i)}^{-1} \\
    &= \frac{1}{k} \sum_{i=1}^n \1_{\set{\KVV_i\in  A\cap L_m}} \kvol\pa{\KVV_i,\cD_n} \\
    &= \frac{1}{k} \sum_{i=1}^n \1_{\set{\KVV_i\in  A\cap L_m}}  \pa{\int_{\R^d}\1_{\cB\pa{\KVV_i,r_k(\KVV_i)}}(\kvv) d \kvv} \\ 
    &= \frac{1}{k} \int_{\R^d} \sum_{i=1}^n \1_{\set{\KVV_i\in  A\cap L_m}}  \1_{\cB\pa{\KVV_i,r_k(\KVV_i)}}(\kvv) d \kvv.
\end{align*}

Let us now introduce, for every $\varepsilon>0$, the event $\Omega_\varepsilon = \set{\max_i r_k(Z_i) \leq \varepsilon}$.
It turns out that on $\Omega_\varepsilon$, every ball with center in $L_m$ is contained in $L_m^\varepsilon:= \set{\kvv\in\R^d, d\pa{\kvv, L_m} \leq \varepsilon} $, that is $\cB\pa{\KVV_i,r_k(\KVV_i)} \subset L_m^\varepsilon $.
This implies that the previous integral over $\R^d$ actually reduces to an integral over $L_m^\varepsilon$, and 
\begin{align*}
    \hat{N}_{k,n}^2  \1_{\Omega_\varepsilon} \leq \frac{1}{k} \int_{L_m^\varepsilon} \sum_{i=1}^n \1_{\cB\pa{\KVV_i,r_k(\KVV_i)}}(\kvv) d \kvv ,
\end{align*}
where the summation in the right-hand side is counting the number of points among $Z_1,\ldots,Z_n$ which are among the $k$ nearest neighbors of $z$.
Stone's lemma \citep[Corrollary 6.1]{gyorfiDistributionFreeTheoryNonparametric2002} then establishes that there exists a constant $\gamma_d>0$ only depending on $d$ such that, almost surely,
\begin{align*}
\sum_{i=1}^n \1_{\cB\pa{\KVV_i,r_k(\KVV_i)}}(\kvv) \leq k \gamma_d .    
\end{align*}
It then results that
\begin{equation}\label{eq:bound_hat_N_small_density}
\hat{N}_{k,n}^2  \1_{\Omega_\varepsilon} \leq \gamma_d \vol\pa{L_m^\varepsilon} .
\end{equation}

\bigskip

It remains to show that, for a fixed $\varepsilon>0$, the probability of the event $\Omega_\varepsilon =\set{\max_i r_k(\KVV_i) \leq \varepsilon}$ tends to $1$.
First, note that since $S$ is compact, 
\[a_\varepsilon := \inf_{z\in S}\P \pa{\KVV_1 \in \cB\pa{\kvv, \varepsilon}} > 0.\]
Besides from the classical connection between the distance of a point to its $k$th nearest neighbor and the binomial probability distribution, it comes that
\begin{align*}
    \P\pa{\max_{1\leq i \leq n} r_k(\KVV_i) > \varepsilon} &\leq \sum_{i=1}^n \P \pa{r_k (\KVV_i) > \varepsilon} \\
    &\leq \sum_{i=1}^n \P \pa{\mathrm{Bin}\pa{n-1, a_\varepsilon} \leq k-1} \qquad \text{(by definition of $a_\varepsilon$)}.
\end{align*}

Let us further use Exercise~2.12 in \citep{boucheron2013concentration} with $\epsilon=p/2$ which gives that
\[
\P\pa{\mathrm{Bin}(n-1,p) \leq \frac{(n-1)p}{2}} \leq \exp\pa{- \frac{(n-1)p}{2}}.
\]
Replacing $p$ with $a_\varepsilon$, one can choose $n$ large enough such that $k - 1 \leq \frac {(n-1)p}{2} = \frac{(n-1) a_\varepsilon}{2}$, which leads to conclude that
\begin{align*}
\P\pa{\max_{1\leq i \leq n} r_k(\KVV_i) > \varepsilon} \leq n \exp\pa{-\frac{(n-1)a_\varepsilon}{2}} \xra{n\to \infty} 0,
\end{align*}
hence $\P(\Omega_\varepsilon) \to 1$, as $n\to +\infty$.

Combining this last result with \cref{eq:bound_hat_N_small_density} assures that, almost surely,
\begin{align}\label{eq.lim.N2.almost.sure}
    \lim_{n\to+\infty} \hat{N}_{k,n}^2 \leq \gamma_d \Vol\pa{L_m^\varepsilon} .
\end{align}

\subsubsection*{Conclusion of the proof.}

Now combining \cref{eq.N1.almost.sure} and \cref{eq.lim.N2.almost.sure}, it comes
\begin{align*}
\lim_{n\to+\infty}\abs{\hat{N}_{k,n} - \Vol\pa{A}} & \leq \lim_{n\to+\infty}\abs{\hat{N}_{k,n}^1 - \Vol\pa{A}} + \abs{\hat{N}_{k,n}^2} \\
&  \leq {\Vol\pa{A\cap L_m}} + \gamma_d \Vol\pa{L_m^\varepsilon} \\
& \leq \Vol\pa{L_m} + \gamma_d \Vol\pa{L_m^\varepsilon}.
\end{align*}

Now, $S$ compact and $\rho$ (uniformly) continuous both imply that 
\[
    \Vol\pa{L_m^\varepsilon} \xra{\varepsilon\downarrow 0} \Vol\pa{L_m},\quad \text{ and that } \quad \Vol\pa{L_m} \xra{m\downarrow 0} 0.
\]
Hence, $\hat{N}_{k,n}$ converges almost surely to $\Vol(A)$.

Since the proof that $\hat{D}_{k,n}$ converges almost surely to $\Vol\pa{S}$ exactly follows the same track, for every Borel set $A\subset S$, one eventually gets that 
\[
    \P\pa{\KVV_I\in A} = \E\bra{\frac{\hat{N}_{k,n}}{\hat{D}_{k,n}}} \xra{n\to\infty} \frac{\Vol(A)}{\Vol(S)},
\] 
which proves the claim.
\end{proof}

\clearpage
\subsection{Additional Numerical Experiments}
\label{app:numerical_experiments}

\subsubsection{Synthetic data}
\label{app:synthetic_data}

\begin{figure}[h]
    \begin{subfigure}[c]{0.49\linewidth}
    \centering
            \includegraphics[width=0.95\linewidth]{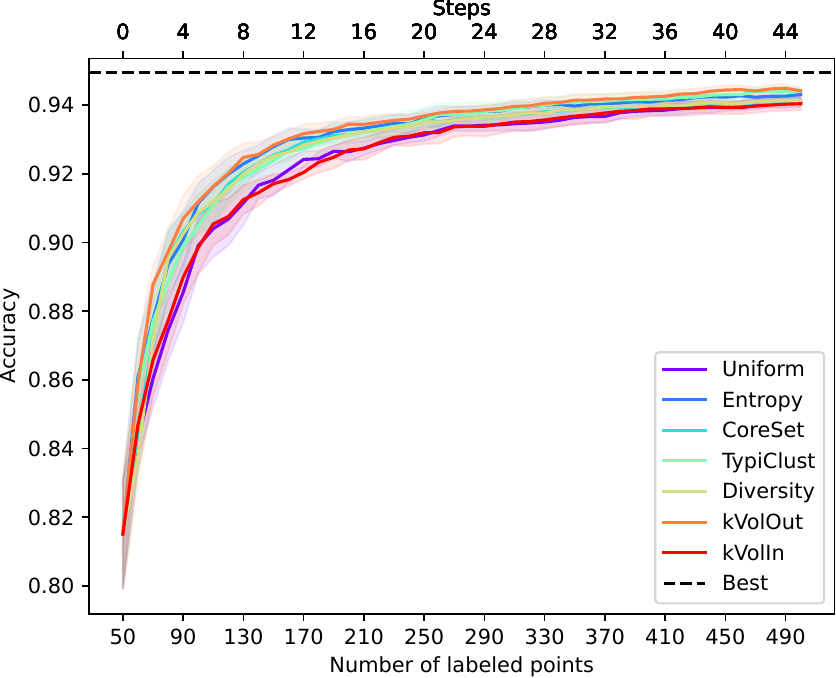}
            \caption{Scenario No. 3 (balanced)}
    \end{subfigure}
    \begin{subfigure}[c]{0.49\linewidth}
    \centering
            \includegraphics[width=0.95\linewidth]{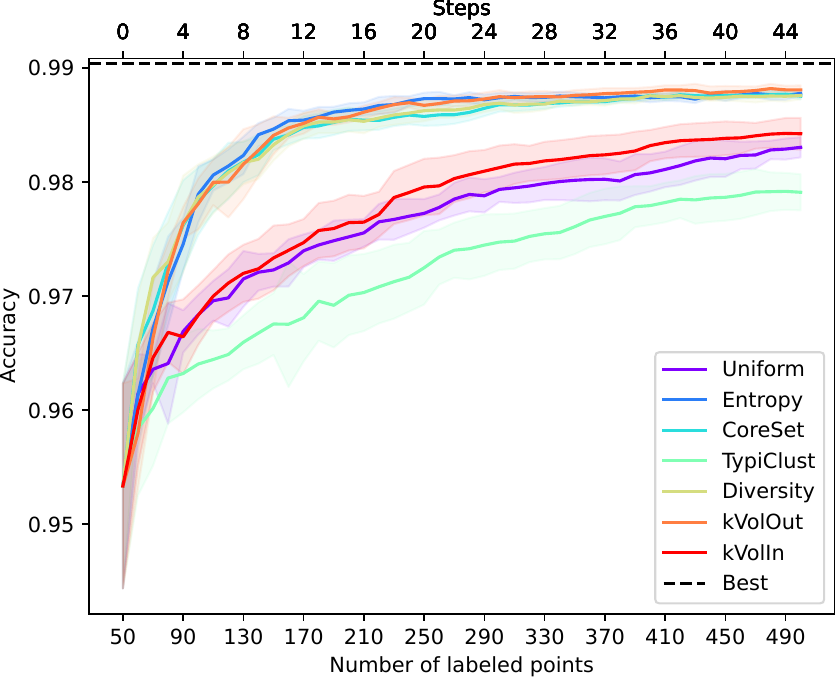}
            \caption{Scenario No. 4 (imbalanced)}
            \label{fig:synthetic_exp_dim20_initbs50_imbalanced}
    \end{subfigure}
    \caption{
    Accuracy comparison during active training. Initial batch size $b_0=50$, step batch size $b=10$ and budget $B=500$. The dimension of the data space is 20.}
    \label{fig:synthetic_exp_dim20_initbs50}
\end{figure}

\begin{figure}[h]
    \centering
    \begin{subfigure}[c]{0.49\linewidth}
            \includegraphics[width=0.95\linewidth]{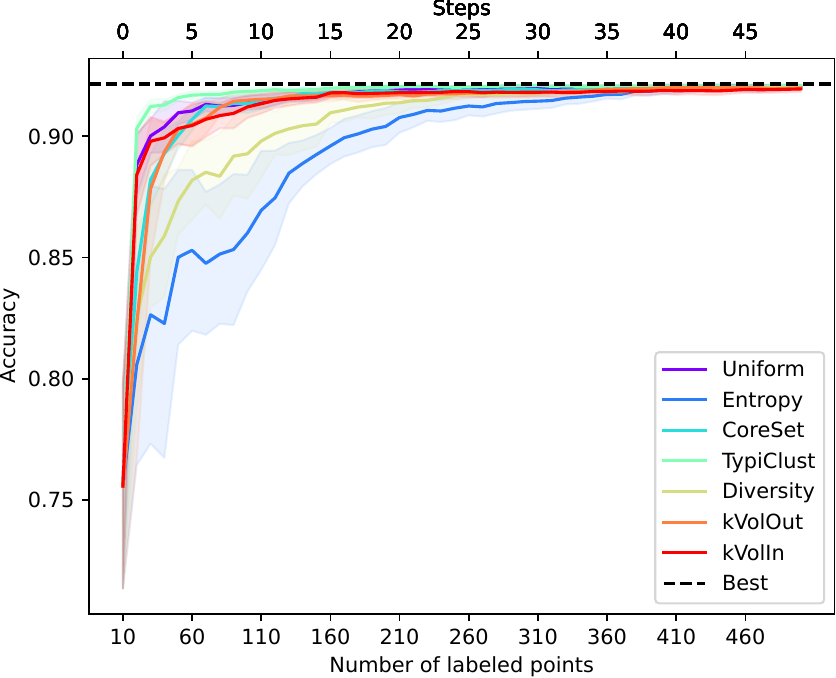}
            \caption{Scenario No. 1}
    \end{subfigure}
    \begin{subfigure}[c]{0.49\linewidth}
            \includegraphics[width=0.95\linewidth]{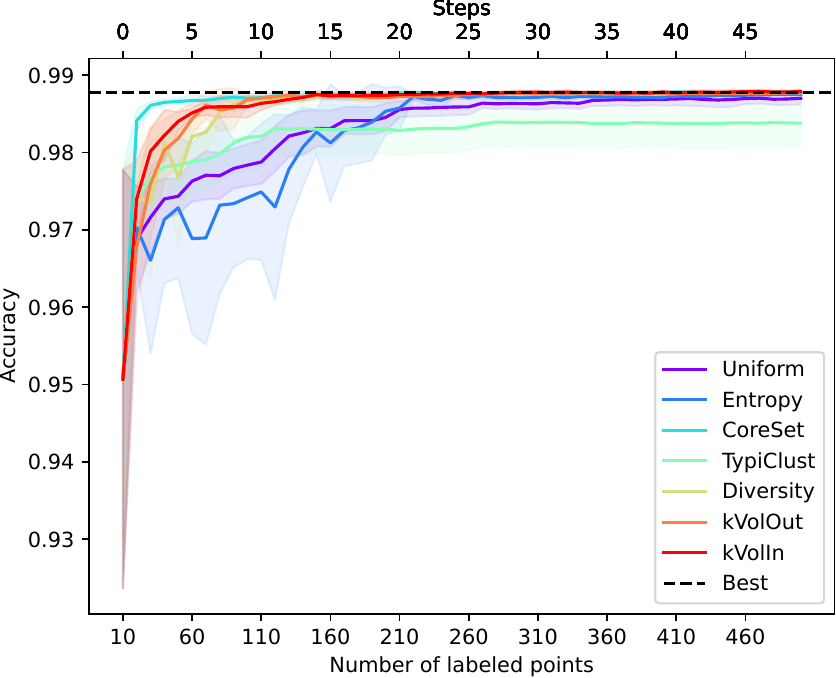}
            \caption{Scenario No. 2}
    \end{subfigure}
    \begin{subfigure}[c]{0.49\linewidth}
            \includegraphics[width=0.95\linewidth]{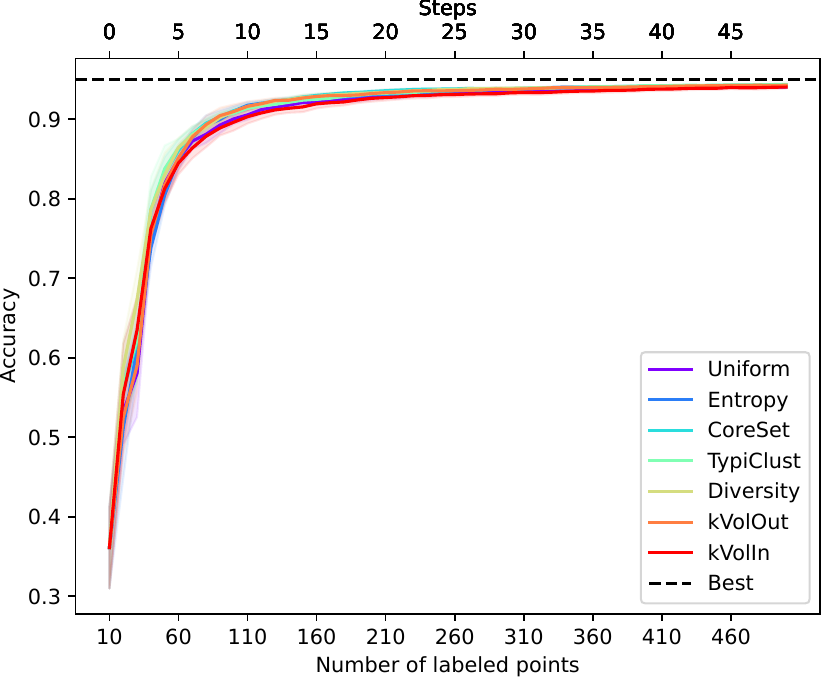}
            \caption{Scenario No. 3}
    \end{subfigure}
    \begin{subfigure}[c]{0.49\linewidth}
            \includegraphics[width=0.95\linewidth]{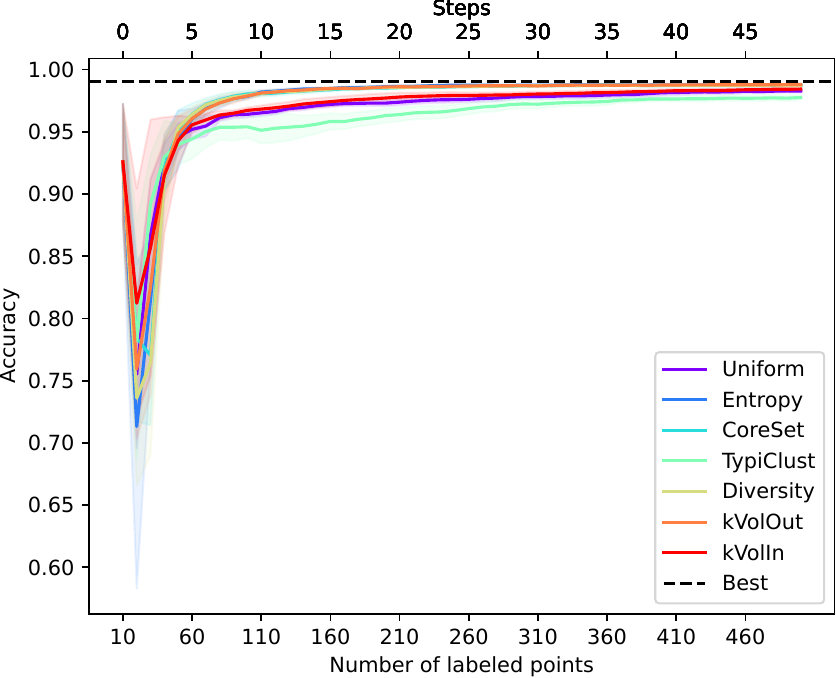}
            \caption{Scenario No. 4}
    \end{subfigure}
    \caption{
    Accuracy comparison during active training. Initial batch size $b_0=10$, step batch size $b=10$ and budget $B=500$.}
\end{figure}

\begin{figure}
    \centering
    \begin{subfigure}[c]{0.49\linewidth}
            \includegraphics[width=0.95\linewidth]{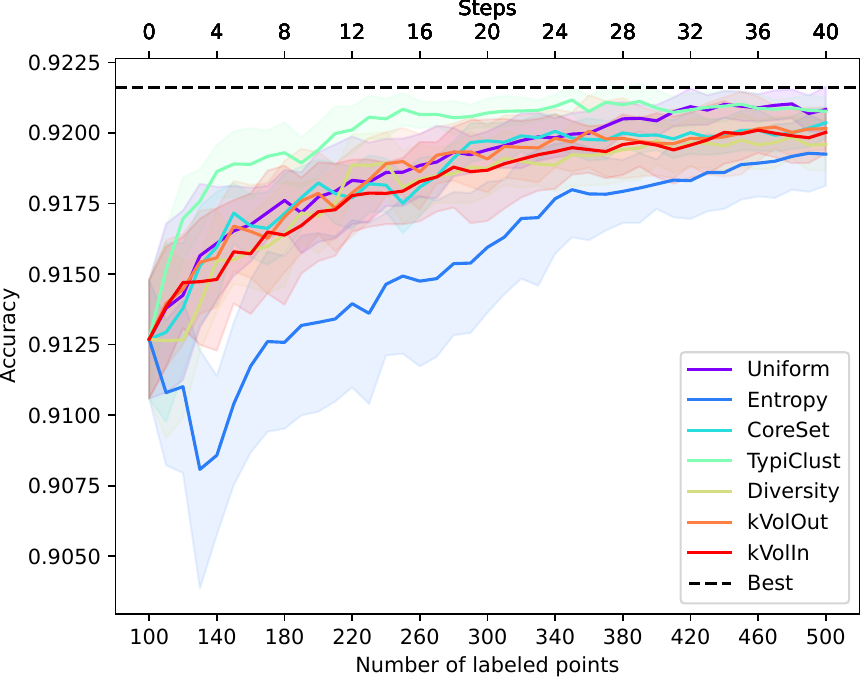}
            \caption{Scenario No. 1}
    \end{subfigure}
    \begin{subfigure}[c]{0.49\linewidth}
            \includegraphics[width=0.95\linewidth]{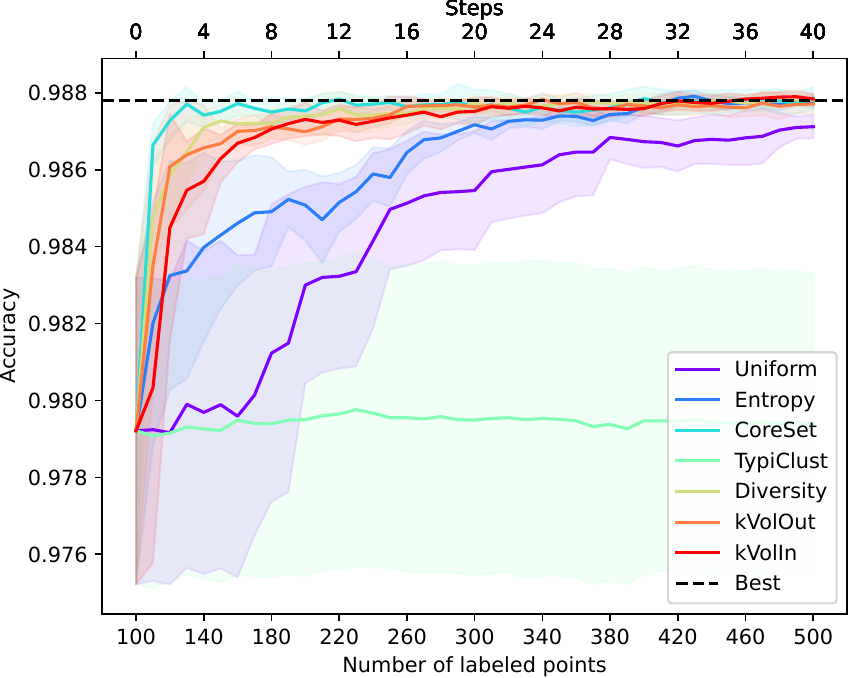}
            \caption{Scenario No. 2}
    \end{subfigure}
    \begin{subfigure}[c]{0.49\linewidth}
            \includegraphics[width=0.95\linewidth]{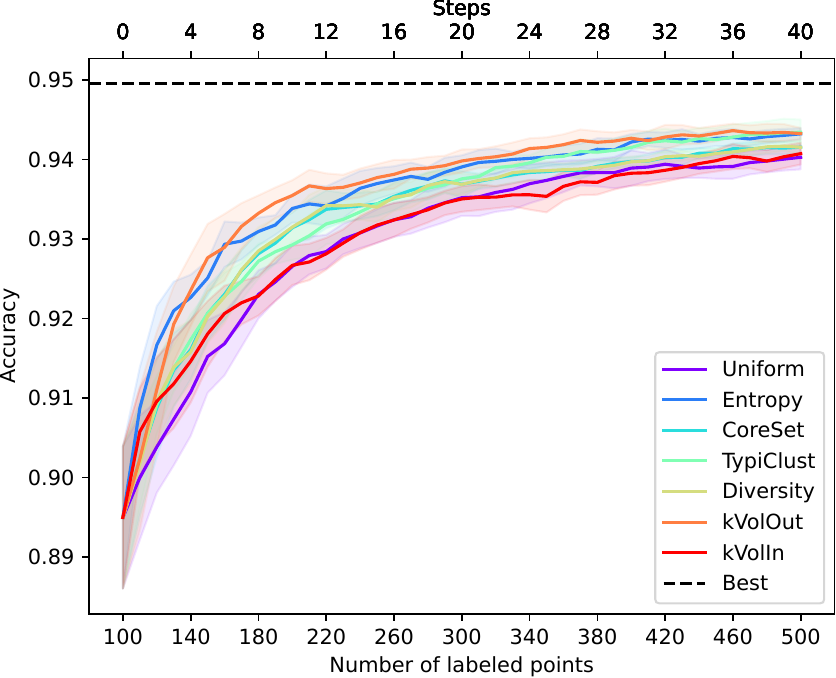}
            \caption{Scenario No. 3}
    \end{subfigure}
    \begin{subfigure}[c]{0.49\linewidth}
            \includegraphics[width=0.95\linewidth]{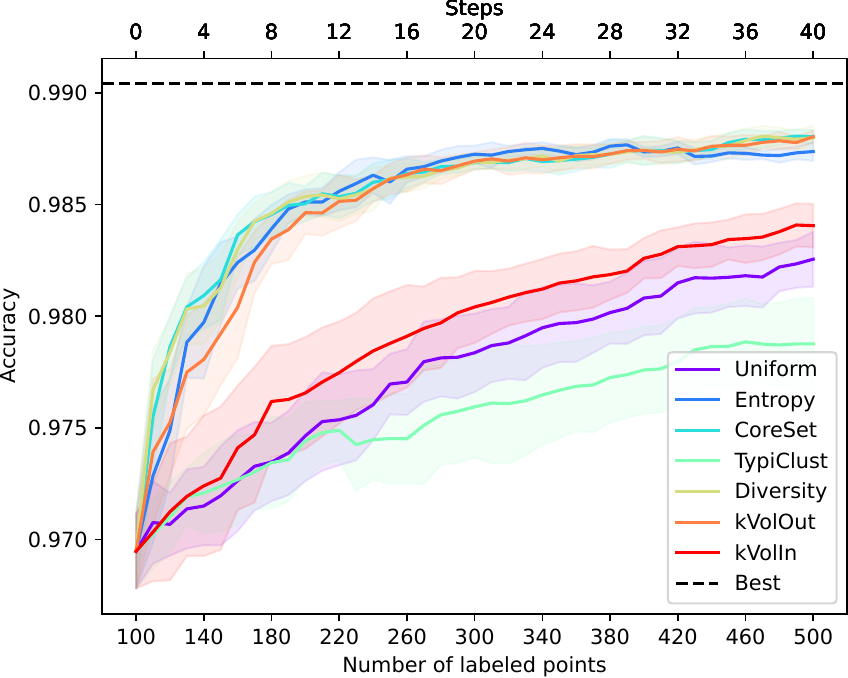}
            \caption{Scenario No. 4}
    \end{subfigure}
    \caption{
    Accuracy comparison during active training. Initial batch size $b_0=100$, step batch size $b=10$ and budget $B=500$.}
\end{figure}

\clearpage
\subsubsection{Tabular data}
\label{app:tabular_data}

\begin{figure}[h]
    \centering
    \begin{subfigure}[c]{0.49\linewidth}
            \includegraphics[width=0.95\linewidth]{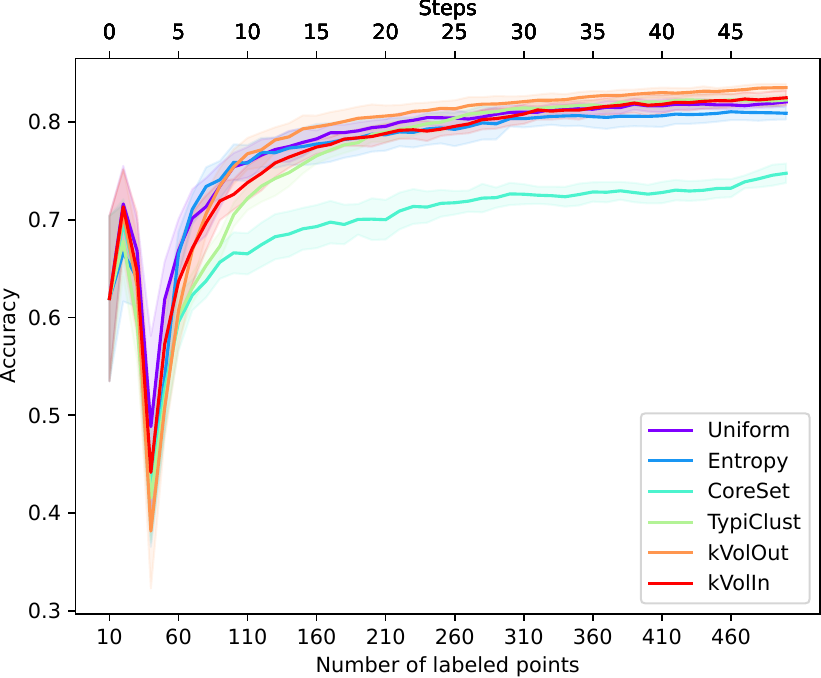}
            \caption{lda.}
    \end{subfigure}
        \begin{subfigure}[c]{0.49\linewidth}
            \includegraphics[width=0.95\linewidth]{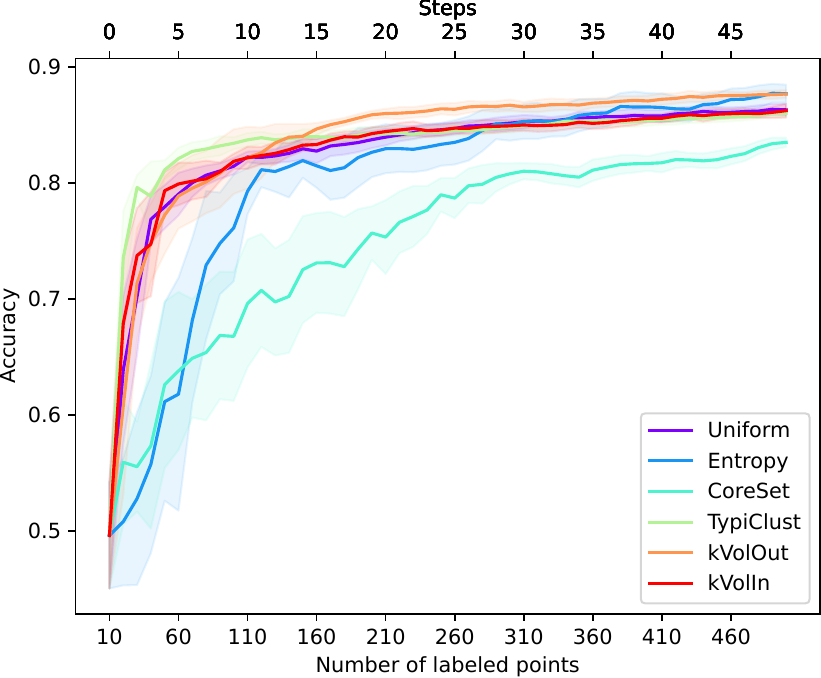}
            \caption{svm.}
    \end{subfigure}
    \begin{subfigure}[c]{0.49\linewidth}
            \includegraphics[width=0.95\linewidth]{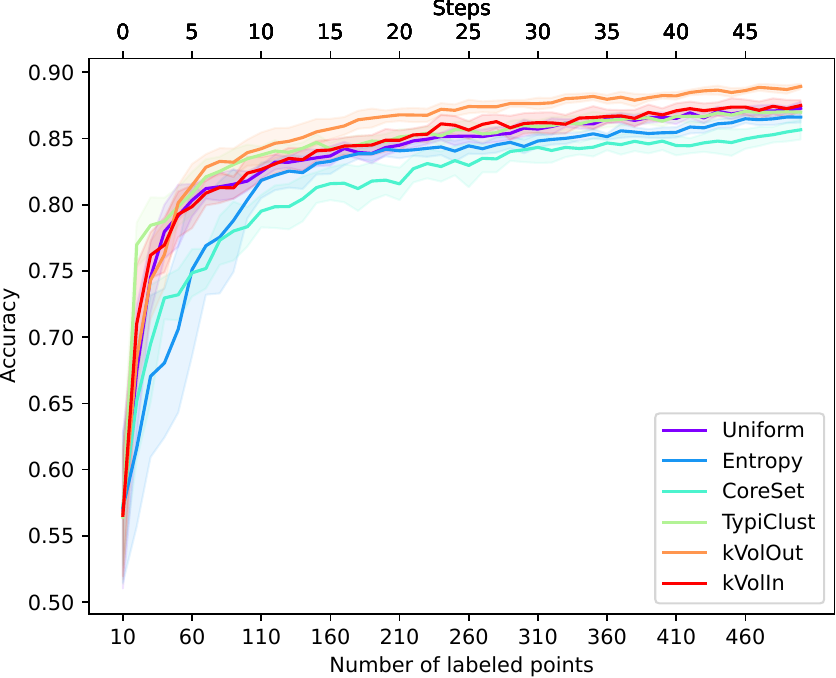}
            \caption{rf.}
    \end{subfigure}
    \caption{Accuracy comparison during active training. Initial batch size $b_0=10$, step batch size $b=10$ and budget $B=500$. The dataset id is 182.}
    \label{fig:tab_data_id182_imbalanced_app}
\end{figure}


\begin{figure}
    \centering
    \begin{subfigure}[c]{0.49\linewidth}
            \includegraphics[width=0.95\linewidth]{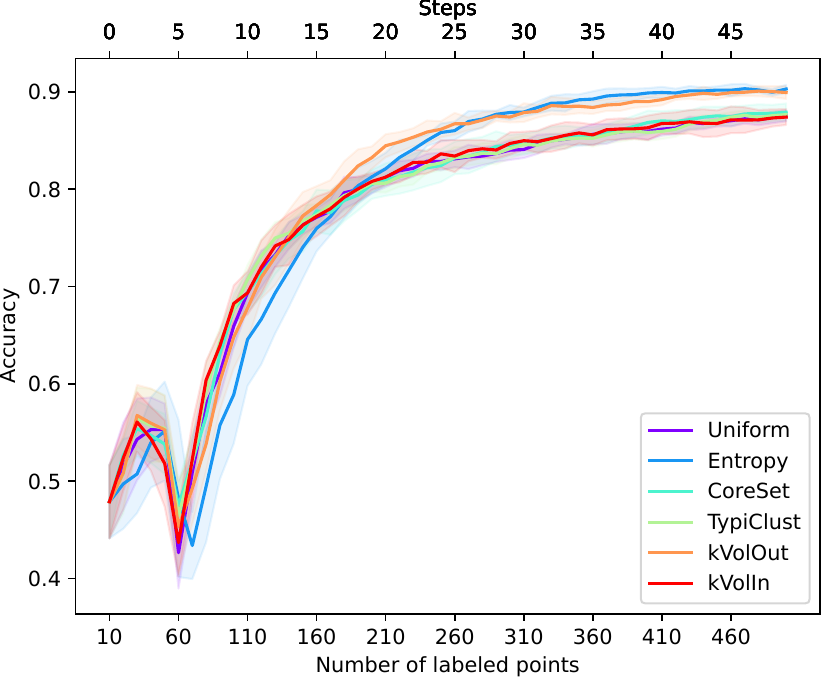}
            \caption{lda.}
    \end{subfigure}
    \begin{subfigure}[c]{0.49\linewidth}
            \includegraphics[width=0.95\linewidth]{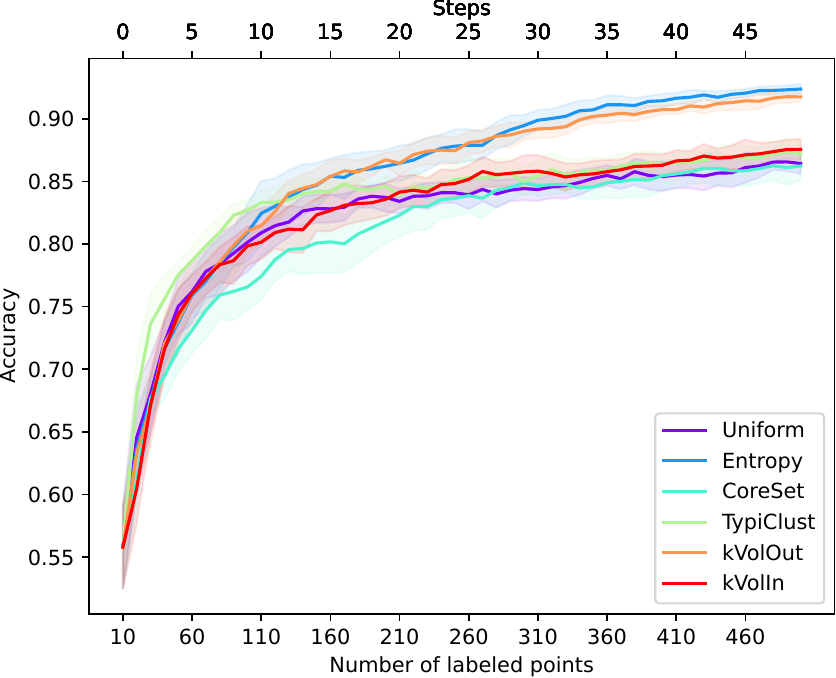}
            \caption{mlr.}
    \end{subfigure}
        \begin{subfigure}[c]{0.49\linewidth}
            \includegraphics[width=0.95\linewidth]{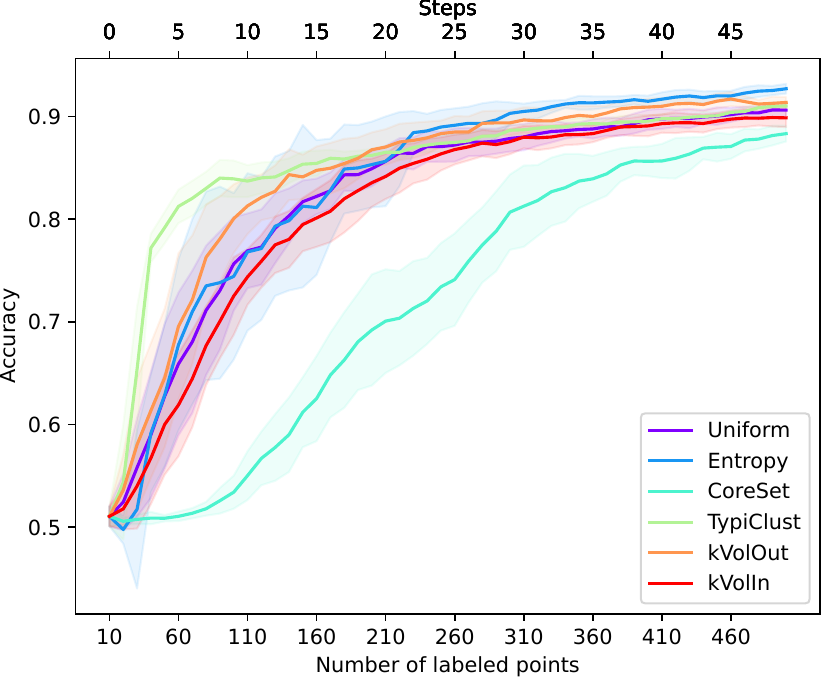}
            \caption{svm.}
    \end{subfigure}
    \begin{subfigure}[c]{0.49\linewidth}
            \includegraphics[width=0.95\linewidth]{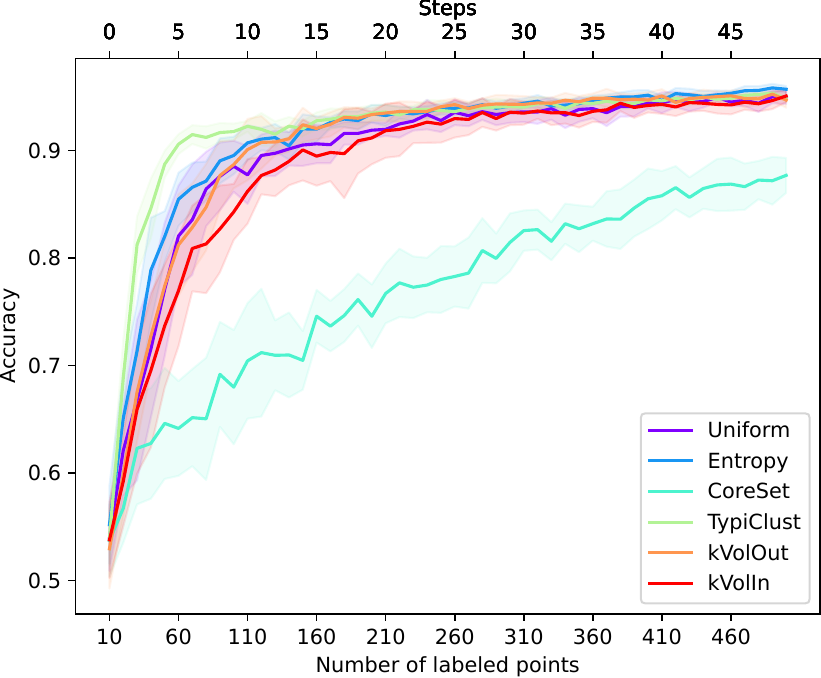}
            \caption{rf.}
    \end{subfigure}
    \begin{subfigure}[c]{0.49\linewidth}
            \includegraphics[width=0.95\linewidth]{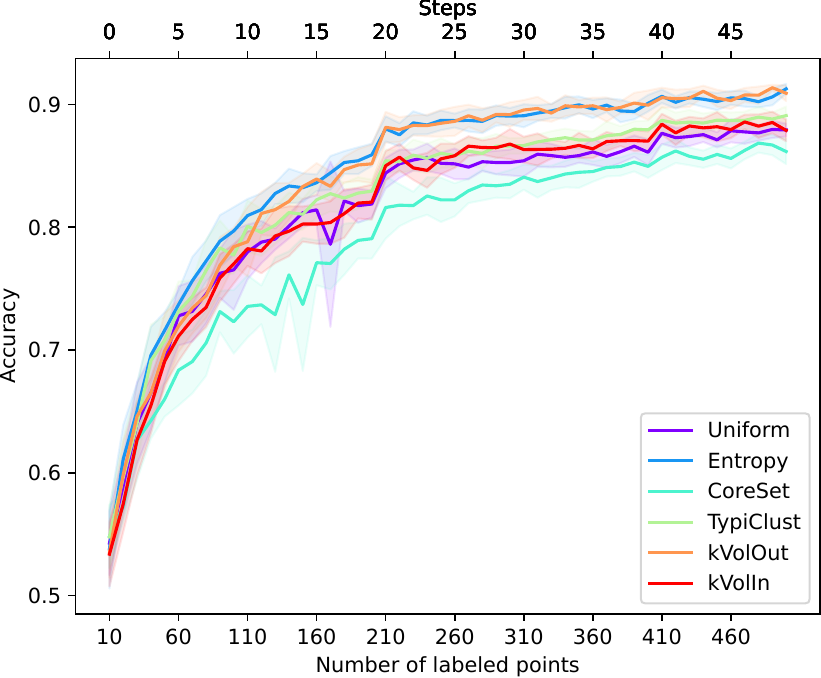}
            \caption{mlp.}
    \end{subfigure}
    \caption{Accuracy comparison during active training. Initial batch size $b_0=10$, step batch size $b=10$ and budget $B=500$. The dataset id is 46.}
    \label{fig:tab_data_id46_imbalanced}
\end{figure}

\begin{figure}[h]
    \centering
    \begin{subfigure}[c]{0.49\linewidth}
            \includegraphics[width=0.95\linewidth]{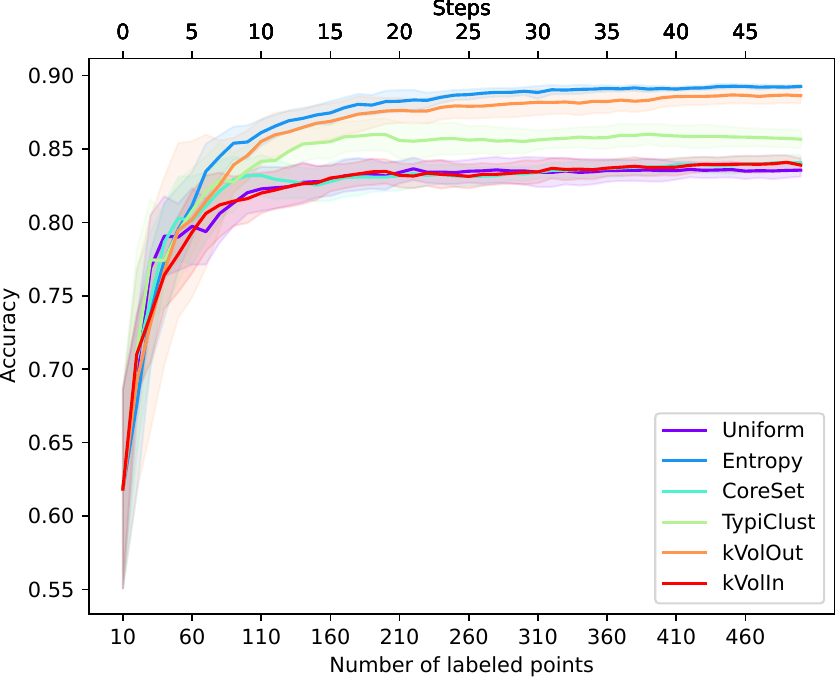}
            \caption{lda.}
    \end{subfigure}
    \begin{subfigure}[c]{0.49\linewidth}
            \includegraphics[width=0.95\linewidth]{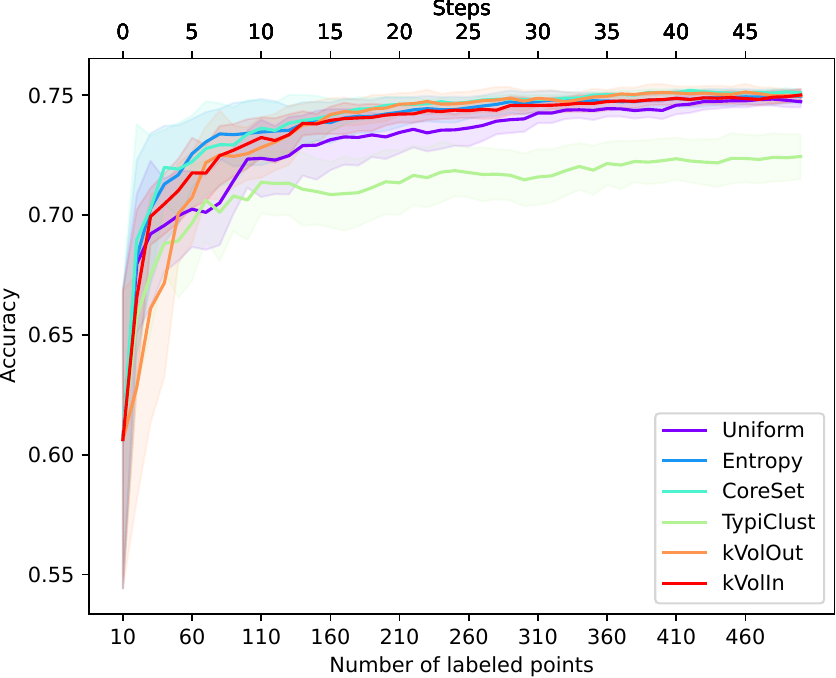}
            \caption{mlr.}
    \end{subfigure}
        \begin{subfigure}[c]{0.49\linewidth}
            \includegraphics[width=0.95\linewidth]{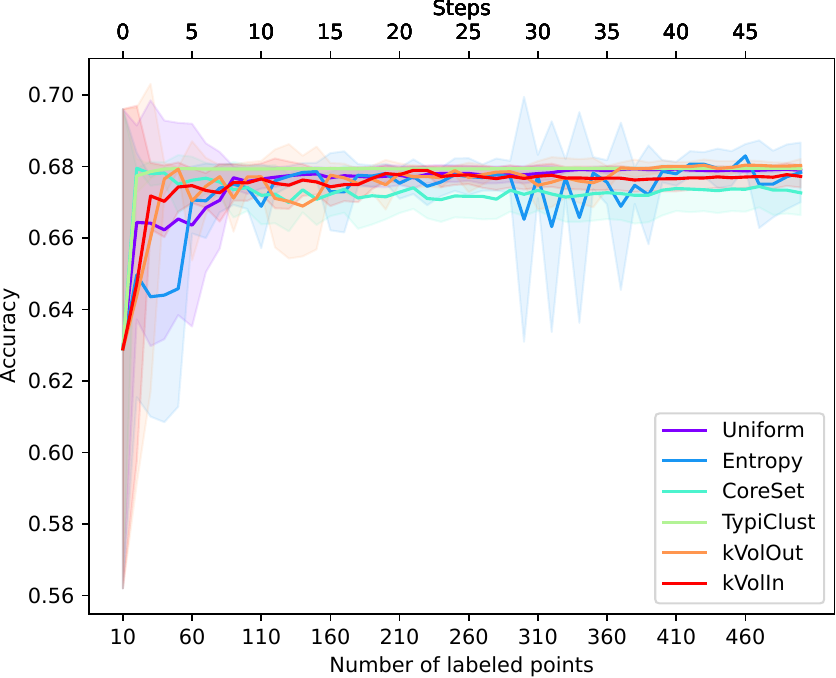}
            \caption{svm.}
    \end{subfigure}
    \begin{subfigure}[c]{0.49\linewidth}
            \includegraphics[width=0.95\linewidth]{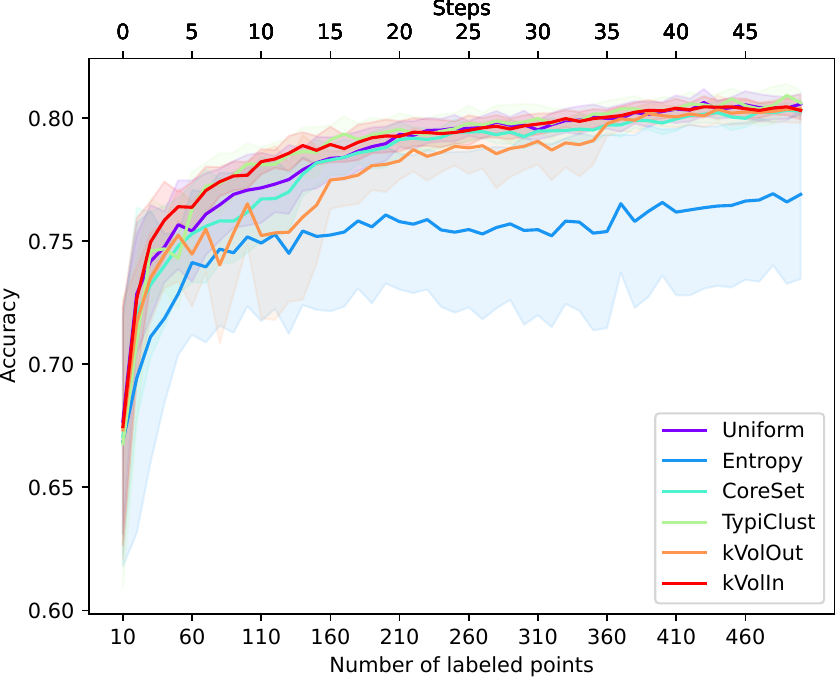}
            \caption{rf.}
    \end{subfigure}
    \begin{subfigure}[c]{0.49\linewidth}
            \includegraphics[width=0.95\linewidth]{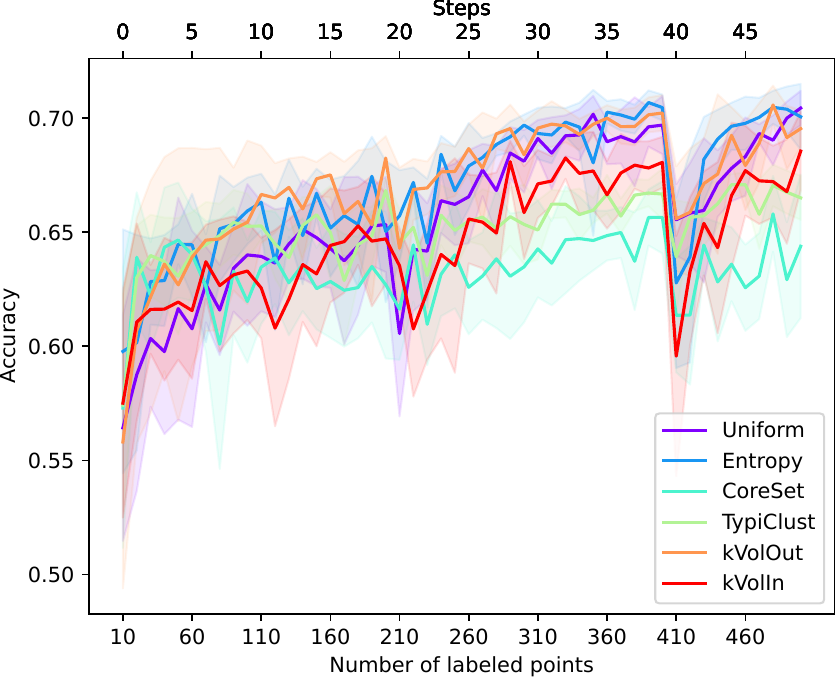}
            \caption{mlp.}
    \end{subfigure}
    \caption{Accuracy comparison during active training. Initial batch size $b_0=10$, step batch size $b=10$ and budget $B=500$. The dataset id is 846.}
\end{figure}

\clearpage
\subsubsection{Image data}
\label{app:image_data}

\begin{figure}[h]
    \centering
    \begin{subfigure}[c]{\linewidth}
    \centering
            \includegraphics[width=0.8\linewidth]{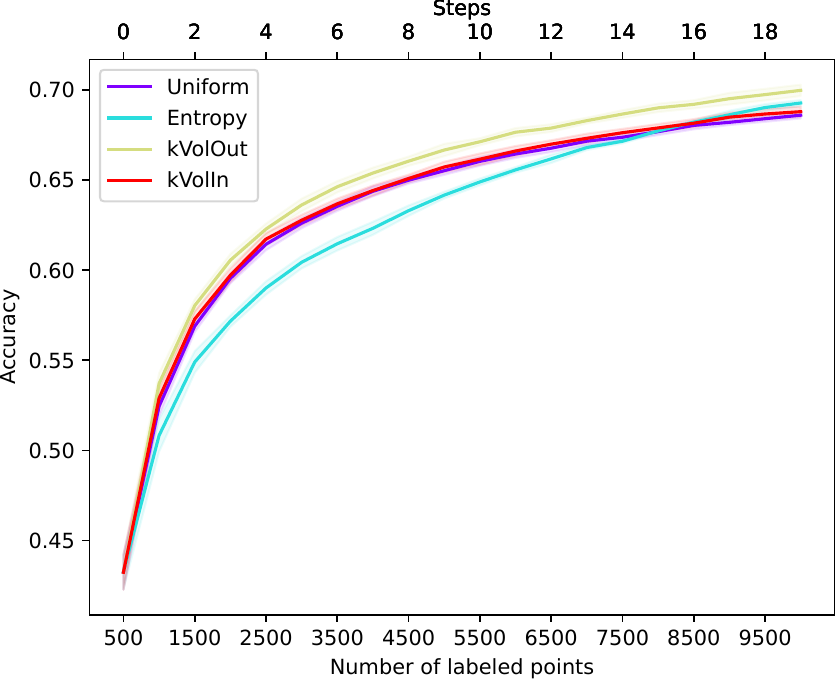}
            \caption{CIFAR100}
    \end{subfigure}
    \caption{Accuracy comparison during active training. Initial batch size $b_0=500$, step batch size $b=500$ and budget $B=10000$.}
    \label{fig:image_data_CIFAR100_10000}
\end{figure}

\clearpage
\subsubsection{Which value for $k$?}
\label{app:value_for_k}

\begin{figure}[h]
    \centering
            \includegraphics[width=0.8\linewidth]{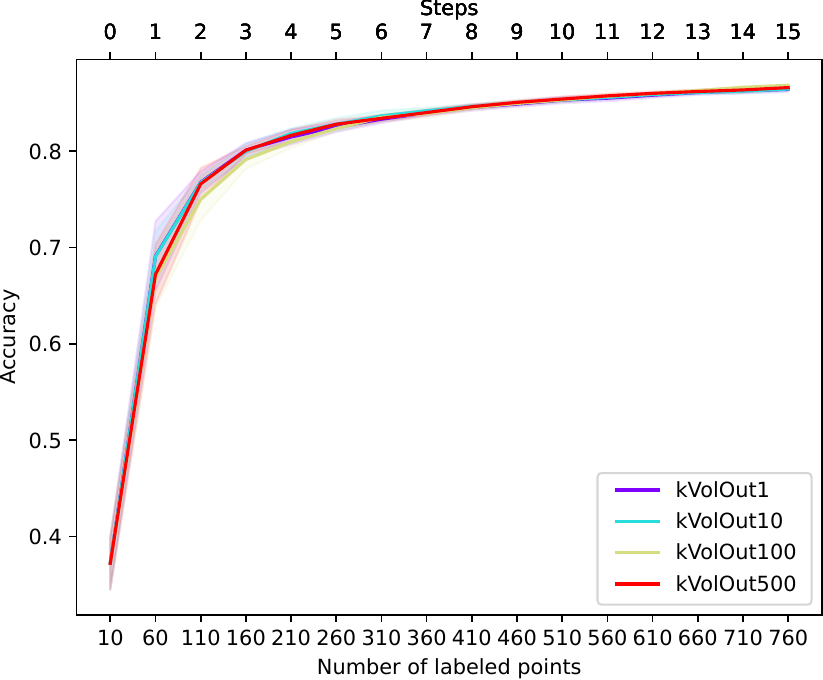}
    \caption{Accuracy comparison during active training for different values of $k$. Initial batch size $b_0=10$, step batch size $b=50$ and budget $B=760$. The dataset is CIFAR10.}
    \label{fig:value_for_k_CIFAR10}
\end{figure}

\subsubsection{Computing time comparison}
\label{app:computing_time_comparison}

The computing time of the different AL strategies tested in this paper is reported in \cref{tab:computing_time_comparison}.
This time is measured for each of the 10 repetitions in the experiments discussed in \cref{sec:num_results,app:numerical_experiments}, and then averaged.
The numerical experiments on image datasets have been performed on an internal cluster on CPU with 48 cores (Intel(R) Xeon(R) Gold 6126 CPU @ 2.60GHz) and 188Go of RAM.

\begin{table}[ht]
    \centering
    \caption{Average computing time per seed (over 10 repetitions) for each strategy (in seconds).}
    \label{tab:computing_time_comparison}
    \begin{tabular}{lcccccc}
    \toprule
          & Uniform & Entropy & CoreSet & TypiClust & kVolOut ($k=4$) & kVolIn ($k=4$) \\
    \midrule
         \cref{fig:image_data_CIFAR100} & 1.70e+03s & 1.71e+03s & 3.06e+03s & 1.07e+04s & 1.73e+03s & 1.90e+03s \\
         \cref{fig:image_data_CIFAR100_10000} & 6.54e+03s & 6.54e+03s & - & - & 6.58e+03s & 6.87e+03s \\
    \bottomrule
    \toprule
        & kVolOut1 & kVolOut10 & kVolOut100 & kVolOut500 & & \\
    \midrule
        \cref{fig:value_for_k_CIFAR10} & 1.14e+02s & 1.37e+02s & 2.01e+02s & 3.43e+02s & & \\
    \bottomrule
    \end{tabular}
\end{table}


\end{document}